\documentclass{article}

\usepackage{microtype}
\usepackage{graphicx}
\usepackage{subfigure}
\usepackage{bbm}
\usepackage{booktabs} % for professional tables

\usepackage{hyperref}

\usepackage[accepted]{icml2024}

\usepackage{amsmath}
\usepackage{amssymb}
\usepackage{mathtools}
\usepackage{amsthm}
\usepackage{xspace}
\usepackage{enumitem}

\usepackage[capitalize,noabbrev]{cleveref}

\definecolor{darkgreen}{HTML}{548235}
\definecolor{darkred}{HTML}{C00000}
\definecolor{darkerblue}{HTML}{240394}
\definecolor{darkblue}{HTML}{2e75B6}
\definecolor{darkyellow}{HTML}{BF9000}
\definecolor{darkpurple}{HTML}{7030A0}

\newcommand{\code}[1]{\texttt{#1}\xspace}
\newcommand{\cmd}[1]{\textcolor{darkerblue}{\textbf{\small{\code{#1}}}}\xspace}

\newcommand{\ours}{LAST\xspace}

\theoremstyle{plain}

\theoremstyle{definition}

\theoremstyle{remark}

\usepackage[textsize=tiny]{todonotes}
\icmltitlerunning{Language-guided Skill Learning with Temporal Variational Inference}

\begin{document}

\twocolumn[
\icmltitle{Language-guided Skill Learning with Temporal Variational Inference}



\icmlsetsymbol{equal}{*}
\icmlsetsymbol{note}{\dag}

\begin{icmlauthorlist}
\icmlauthor{Haotian Fu}{xxx}%,note}
\icmlauthor{Pratyusha Sharma}{yyy}%,note}
\icmlauthor{Elias Stengel-Eskin}{zzz}%,note}
\icmlauthor{George Konidaris}{xxx}
\icmlauthor{Nicolas Le Roux}{ccc,ttt}
\icmlauthor{Marc-Alexandre Côté
}{equal,ttt}
\icmlauthor{Xingdi Yuan}{equal,ttt}
\end{icmlauthorlist}

\icmlaffiliation{xxx}{Brown University}
\icmlaffiliation{yyy}{MIT}
\icmlaffiliation{zzz}{University of North Carolina, Chapel Hill}
\icmlaffiliation{ccc}{MILA}
\icmlaffiliation{ttt}{Microsoft Research}

\icmlcorrespondingauthor{Haotian Fu}{hfu7@cs.brown.edu}

\icmlkeywords{Machine Learning, ICML}

\vskip 0.3in
]



\printAffiliationsAndNotice{* Equal Advising}%\dag Work partially done while interning at MSR}% otherwise use the standard text.

\begin{abstract}
We present an algorithm for skill discovery from expert demonstrations. The algorithm first utilizes Large Language Models (LLMs) to propose an initial segmentation of the trajectories. Following that, a hierarchical variational inference framework incorporates the LLM-generated segmentation information to discover reusable skills by merging trajectory segments. To further control the trade-off between compression and reusability, we introduce a novel auxiliary objective based on the Minimum Description Length principle that helps guide this skill discovery process. 
% We test our framework on BabyAI, a grid world navigation environment, as well as ALFRED, a household simulation environment.
% Our results demonstrate that agents equipped with our method can discover skills that help accelerate learning and outperform baseline skill learning approaches on new long-horizon tasks.
Our results demonstrate that agents equipped with our method are able to discover skills that help accelerate learning and outperform baseline skill learning approaches on new long-horizon tasks in BabyAI, a grid world navigation environment, as well as ALFRED, a household simulation environment.\footnote{Code and videos can be found at \url{https://language-skill-discovery.github.io/}.}
\end{abstract}

\section{Introduction}
\label{intro}

A major issue that makes Reinforcement Learning (RL) hard to use for long-horizon interaction tasks is its sample inefficiency. Conventional Deep RL algorithms explore and learn task-specific policies from scratch, which can require over 10M samples to train just one Atari game~\citep{DBLP:journals/nature/MnihKSRVBGRFOPB15,DBLP:conf/aaai/HesselMHSODHPAS18}. In contrast, humans can play well after just 20 episodes. Humans have a strong set of priors that helps us efficiently adapt to new tasks. Some recent work has made attempts to introduce such priors for embodied agents as well, like constructing world-models~\citep{DBLP:journals/corr/abs-2301-04104, DBLP:journals/corr/abs-2310-16828}, a goal/reward-conditioned policy~\citep{DBLP:journals/tmlr/ReedZPCNBGSKSEBREHCHVBF22, DBLP:conf/nips/LeeNYLFGFXJMM22}, or a toolbox of extracted skills/options~\citep{DBLP:conf/corl/PertschLL20, pmlr-v162-liu22t, DBLP:journals/corr/abs-2305-16291}. Skills are temporally-extended policies~\citep{DBLP:journals/ai/SuttonPS99}; each skill is executed for a certain number of steps until switching to another skill. While the observation that pre-specifying the set of skills enables effective generalization to longer-horizon tasks, an effective algorithm to autonomously discover the set of domain-specific skills to enable long-horizon planning remains challenging. Our goal is to enable the discovery of reusable skills from a dataset of expert demonstrations (i.e., trajectories) of an agent performing various complex tasks, and use these skills to solve new complex tasks more efficiently. 
% We do this by segmenting  trajectories and merging subsequences into skills so that we can use them to replace the primitive low-level actions and solve new complex tasks more efficiently.

Many recent approaches learn skills from expert demonstrations by first slicing the trajectories into segments and merging them with a latent skill representation, and then reconstruct the trajectories from these latent variables~\citep{DBLP:conf/icml/Shankar020, DBLP:conf/iclr/ShankarTP020, DBLP:conf/icml/KipfLDZSGKB19}. Such methods in practice often fall into two categories of ``local optima'': 1. every single step in one trajectory is segmented as one skill and the learned skill space is like a continuous resemblance of the original action space. 2. the agent directly marks the whole trajectory as one skill and does not do any segmentation. In the first case, it is hard to accelerate learning on new tasks as the skills are basically the original actions. For the second case, the learned skills may perfectly reconstruct the training trajectories but cannot generalize on a new task as long as it is different from the training ones.
% While the skills are reusable in this case, they cannot help us solve new tasks faster as they do not shorten the task horizon. This also makes the skills hard to transfer since a whole trajectory is not reusable enough. 

\begin{figure}[t!]
\centering
\includegraphics[width=0.9\linewidth]%{icml2024intro.pdf}
{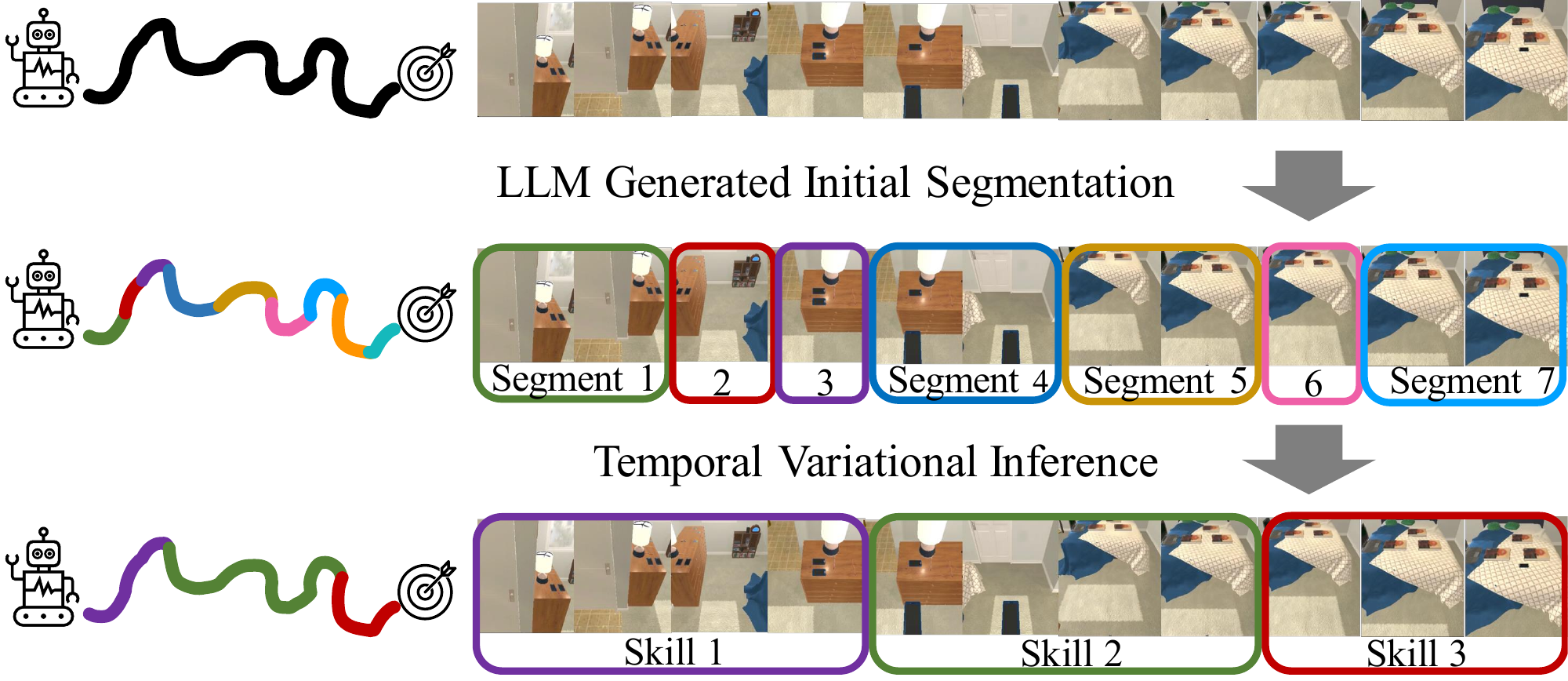}
\vspace{-0.5em}
    %\caption{Illustration of our two-step trajectory segmentation and merging for learning skills.} 
    \caption{The trajectory segmentation and merging procedure.} 
    \vspace{-1em}
    \label{fig:intro}
\end{figure}

\begin{figure*}[htbp]
\centering
    \includegraphics[width=0.39\linewidth]{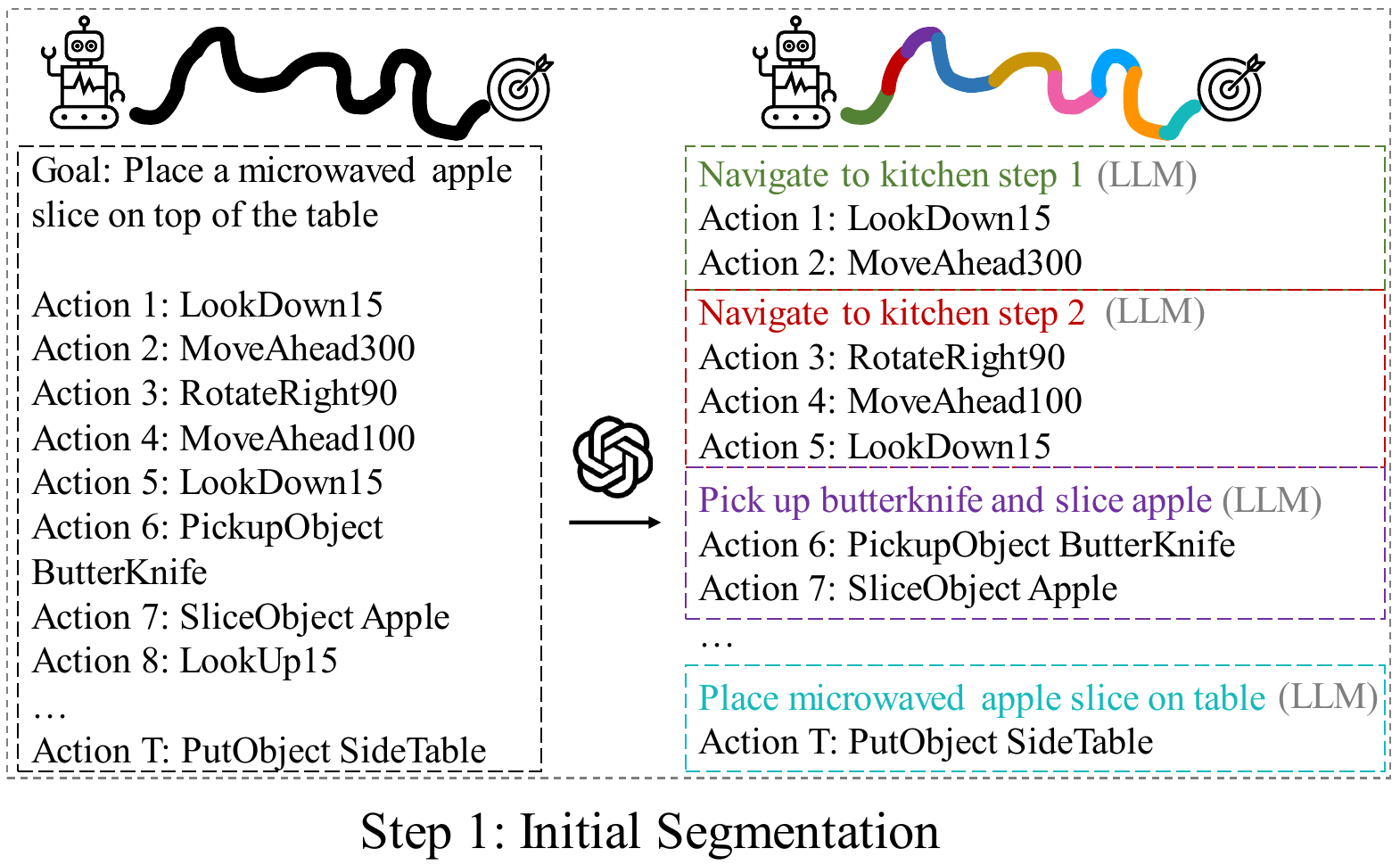}
    \includegraphics[width=0.36\linewidth]{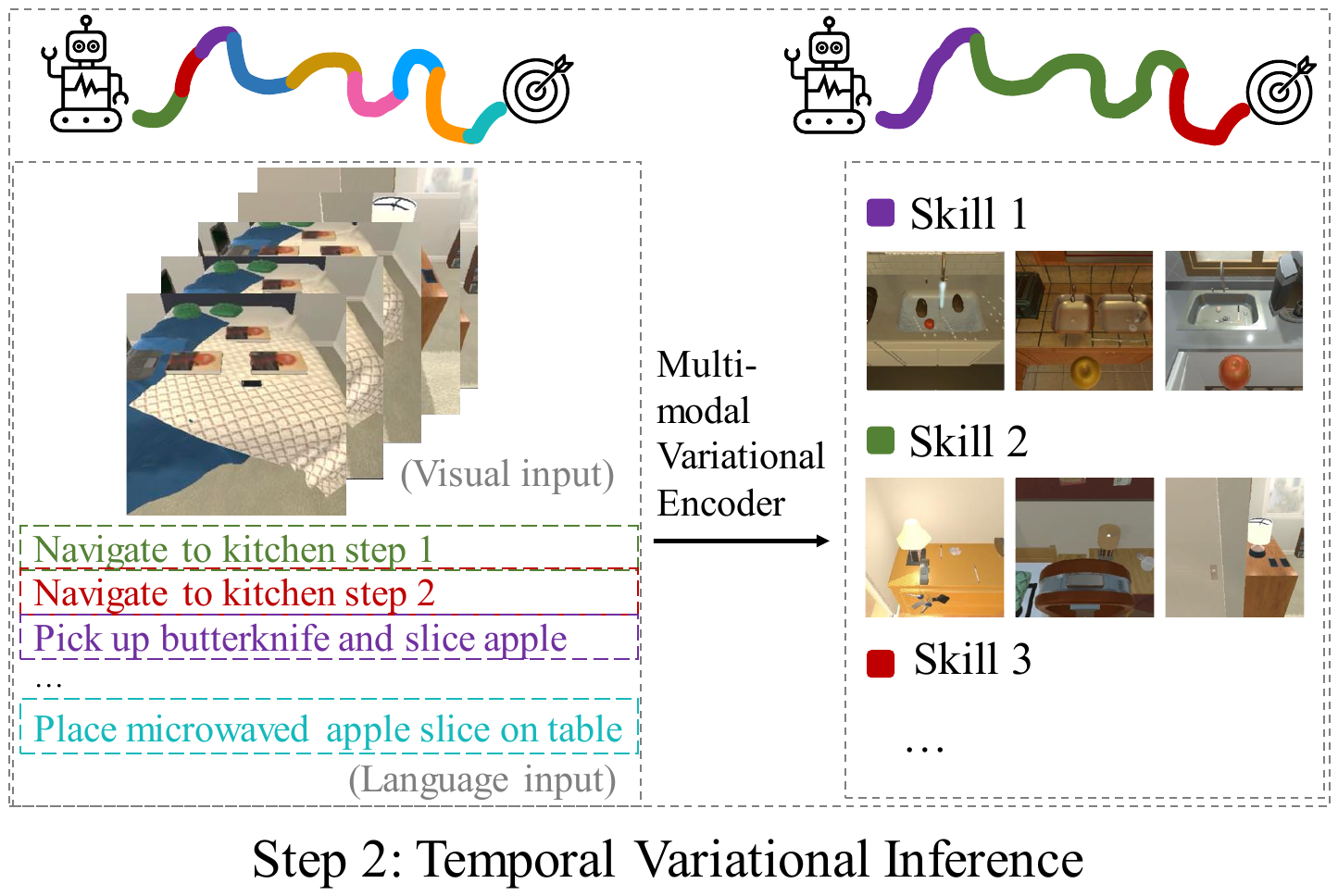}
    \includegraphics[width=0.215\linewidth]{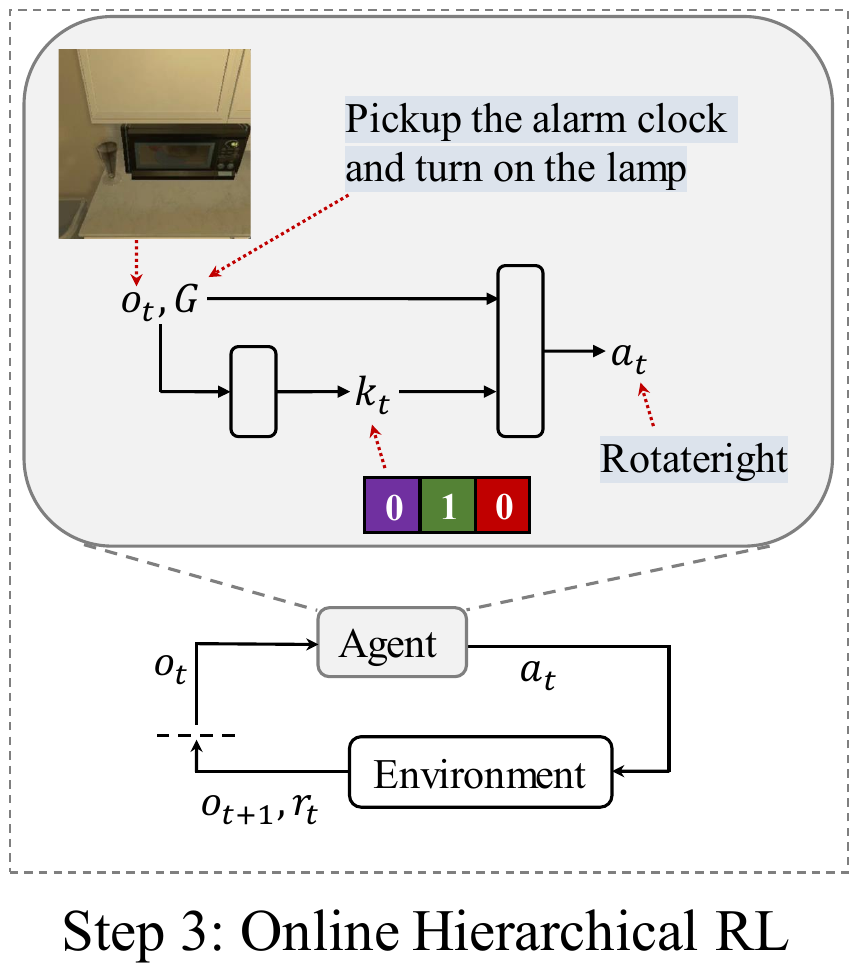}
    %\vspace{-0.8em}
    \caption{Overall framework of {\bf \ours}. \textbf{Step 1}: given a dataset of expert demonstrations, we query an LLM (only using the goal and actions as input) for an initial segmentation and a language description for each segment. \textbf{Step 2}: temporal variational inference takes in multi-modal data as input to improve upon the segmentation by merging different subsequences into skills. \textbf{Step 3}: online hierarchical RL on new tasks leveraging the learned skills which can greatly shorten the task horizon and help the agent efficiently learn on new tasks.} 
    %\vspace{-1em}
    \label{fig:general}
\end{figure*}
Why is it hard to balance between compression and reusability 
when learning from long trajectories? We believe that this is because the search space for trajectory segmentation is too large especially when the task horizon is long. The number of ways to segment a sequence grows exponentially with respect to the horizon, which will make the learning difficult and prone to poor local optima. 
% Training with information theoretical objectives~\citep{DBLP:conf/nips/JiangLEKF22} can make this search space smaller, but when the horizon is long it still admits many different solutions, of which many can not produce reusable skills. 
We propose to narrow down the search space via {\bf Initial Segmentation} and  {\bf Language-Augmented Temporal Variational Inference}. As shown in Fig.~\ref{fig:intro}, for initial segmentation, we use an LLM to segment each trajectory into many short subsequences and generate corresponding language annotations. Then, we propose a temporal variational inference framework that can improve upon the initial segmentation by merging short subsequences into longer ones and integrate them into reusable skills. The search space is gradually narrowed down via 1) forcing the agent to never split a segment generated by the LLM---only merge them into larger ones; 2) maximizing the trajectory reconstruction likelihood and choosing skills that provide the shortest descriptions of the trajectory. The two steps together greatly reduce the number of possible ways to segment a sequence while still maintaining the reusability of the learned skills.
% In addition, our method can perform faster online adaptation on new tasks without additional language instructions.

Specifically, we propose a novel algorithm that can effectively merge semantic priors from language models with temporal variational inference to discover skills. Our contributions are: 1) We propose a method that leverages a pretrained LLM to generate an initial segmentation of the expert trajectories, which greatly decreases the search space for skill segmentation and provides additional learning signals for the following skill discovery steps. 2) We propose a hierarchical variational inference framework that can incorporate the generated language supervision and discover skills on top of initial trajectory segmentations. 3) We augment the temporal variational inference process with a novel auxiliary training objective following Minimum Description Length Principle, which further helps compression and discover reusable skills. 4) The proposed online hierarchical RL framework enables the agent to quickly adapt to new long-horizon tasks with the learned skills in BabyAI~\citep{DBLP:conf/iclr/Chevalier-Boisvert19} and ALFRED~\citep{DBLP:conf/cvpr/ShridharTGBHMZF20}, the latter of which is a highly complex household simulation environment with multimodal inputs and long-horizon tasks.
%\footnote{Demo videos are available at \href{https://language-skill-discovery.github.io/}{language-skill-discovery.github.io}.}

\section{Related Work}
\label{relatedw}
The proposed approach improves over previous work in two ways: First, methods that use language/language model priors, inadvertently restrict the representation of the skills to natural language alone---this restriction doesn’t apply to our framework. Second, most methods that use variational inference to discover skills from demonstration do not utilize any external semantic guidance from language models, which makes the skill discovery procedure prone to overfitting to short-horizon plans and the model fails to learn skills that generalize well.

\textbf{Skill learning from demonstrations:} We consider the problem of learning skills from demonstrations~\citep{DBLP:conf/iros/NiekumOKB12, DBLP:conf/rss/NiekumCBMO13, DBLP:conf/iros/MeierTSS11, DBLP:conf/iclr/SharmaSRK19, DBLP:conf/corl/KrishnanFSG17, DBLP:conf/icra/MuraliGKPADG16, DBLP:journals/corr/FoxKSG17}, for tasks where the action space is made of strings.~\citet{DBLP:conf/icml/Shankar020} propose a framework to use Temporal Variational Inference to learn skills from demonstrations. However, it is limited to low-dimensional state space and~\citet{DBLP:conf/nips/JiangLEKF22} shows that such methods~\citep{DBLP:conf/nips/KimAB19} are under-specified and can fail to learn reusable skills. 
% Despite having the information theoretical training objectives, we find empirically that Temporal Variational Inference tends to mark the whole trajectory as a single skill given tasks that are harder and with longer horizon.
% This motivates us to equip our agents with extra supervision and prior knowledge.
Additionally, they propose an auxiliary training objective that encourages compression. However, we find that in practice when the tasks are much harder and require long-horizon reasoning, despite this training objective the agent may still mark the whole trajectory as one skill and does not do any segmentation. In this paper, we argue that besides information theoretical training objectives, more supervision / prior knowledge needs to be considered to learn reusable skills for complex long-horizon planning tasks. 
Methods for learning skills from multiple tasks ~\citep{DBLP:journals/corr/HeessWTLRS16, DBLP:conf/icml/RiedmillerHLNDW18, DBLP:conf/iclr/HausmanS0HR18, DBLP:conf/icml/FuYTL023} and learning to compose the said skills have also been discussed by a rich body of literature~\citep{DBLP:conf/nips/KonidarisB09, DBLP:journals/ijrr/KonidarisKGB12, DBLP:conf/icml/KipfLDZSGKB19, DBLP:conf/icml/BagariaS021}. Recent works also consider skill learning from offline trajectories~\citep{DBLP:conf/iclr/AjayKALN21, DBLP:conf/iclr/RaoSHWZVTAMHH22, DBLP:conf/corl/PertschLL20, DBLP:conf/nips/XuVS22}, containing a large amount of sub-optimal data. In this work, we employ a two-level hierarchical RL framework with a frozen low-level policy.

\textbf{Language and interaction:}
Several studies have focused on utilizing natural language and language models to assist with planning for long-horizon tasks. While some studies have used natural language to represent intermediate planning steps or skills \citep{branavan-etal-2009-reinforcement,mooney,NIPS2013_7cce53cf,pmlr-v70-andreas17a}, others have used language to map directly to sequences of abstract actions \citep{mooney,Tellex2011UnderstandingNL,Misra2017MappingIA,anderson,alfworld} or plan over a known domain model of the environment \citep{song2023llmplanner,arora2023learning,silver2024generalized}. Several of these approaches are also equipped with the use of language models as a distribution over valid sequences of skills \citep{DBLP:conf/acl/Sharma0A22,progprompt,liu2023llmp} and actions themselves \citep{huang2022language}. However, all of these methods require either specifying the planning domain, the set of skills, and/or precise abstract action steps in advance. They do not construct the skill set itself, a crucial aspect of our proposal. 

Recent studies have proposed using natural language to guide the discovery of reusable skills \citep{DBLP:conf/acl/Sharma0A22} from unparsed demonstrations. However, they have assumed that the skills themselves must be represented with natural language 
% and use crowdsourced natural language instructions to guide the discovery
. Although powerful in ``realistic'' domains, natural language as the choice of representation can become restrictive when
%in domains where 
the optimal skill set is hard to specify using natural language alone. In this paper, we use language (and language models) to guide the discovery of skills, while lifting the restriction.
% that the skills themselves need to be represented in natural language 
% Furthermore, we do not utilize any additional human-annotated data. 
BOSS~\citep{DBLP:conf/corl/ZhangZPLRCSL23} starts from a pretrained initial skill library and focuses on learning new skills through online interactions by chaining the existing skills which are guided by LLM directly. SPRINT~\citep{DBLP:journals/corr/abs-2306-11886} proposes to use LLMs to relabel robot trajectories and leverage offline reinforcement learning methods to do skill chaining. Other recent work simultaneously propose the domain model and learn libraries of action operators defined as code-based policies guided by LLMs \cite{DBLP:journals/corr/abs-2305-16291,liu2023llmp,wong2023learning} as a way to learn hierarchical action representations. However, these approaches require a planner and multiple rounds of interaction in the environment thereafter to iteratively verify the correctness and extensibility of the inferred skills.

%\section{Method}
\section{Problem Setup}

We consider learning problems where the agent needs to use the experience from expert demonstrations to quickly solve new RL tasks. We formalize these problems as Partially Observable Markov Decision Processes (POMDPs), defined by a tuple $(S,A,O,\Omega, \mathcal{T},R)$. We use $S$ to denote the state space, $A$ for the action space, $O$ for the observation space, $\Omega: S\rightarrow O$ for the observation function, $\mathcal{T}: S\times A \rightarrow S$ for the transition function, and $R$ as the reward function. The dataset is a set of $M$ goal-conditioned trajectories $\mathcal{D}=\{\tau_i\}_{i=1}^{M}$, and $\tau_i = \{G_i, o_{i1}, a_{i1}, \cdots, o_{iT_i}\}$, where $G$ denotes the task goal, $o \in O$ denotes the observation, $a \in A$ denotes the actions, and $T$ is the length of the trajectory. We assume every action has a semantic meaning described by language. The trajectories are collected from \emph{multiple} tasks so each trajectory $\tau_i$ may have a different reward function.

% New version
In this work, we assume that our policies are mixtures of time-limited, semantically meaningful, sub-policies which we call \emph{skills}, and that these skills will be shared across tasks. Formally, we introduce two new time-dependent variables: $k_t$ is the skill used at time $t$ and $\beta_t$ is a \emph{skill-switching variable} meant to identify when we move from one skill to another, i.e. $k_t = k_{t-1}$ when $\beta_t =0$. 
% The dynamics model over these new variables is defined by $\pi(\beta_t, k_t \mid  o_t) \mid $
% watch out, I call it \beta and not \beta^new as i want to keep notation simple.
 Denoting $\phi = \{\beta_t, k_t\}_{t=1}^T$, we have $p(\tau) = \sum_\phi p(\phi) p(\tau \mid  \phi)$ as joint probability over trajectory and skills. $p(\tau \mid  \phi)$ is the skill (and goal) conditioned policy, and $p(\phi)$ is the high-level policy over the skills. Correctly approximating the posterior distribution $p(\phi \mid  \tau)$ over skills given the trajectories will be critical to extract the skills which we will be able to reuse for new tasks.
%  The generative process can be written as:
%  \begin{equation}
% \label{bgeqa:p}
% \small
% \begin{aligned}
%      &p(\tau, \phi) = p(o_{:T}, a_{:T}, \beta_{:T}, k_{:T}, G) = 
%      p(o_1)p(G)
%      \textstyle \\ 
%      &\prod_{t=1}^{T} p(o_{t+1}\mid o_t, a_t)\pi(a_t\mid o_{:t}, k_{t}, G)
%     p({\beta}_t, k_t\mid o_{:t}, a_{:t-1}, \beta_{:t-1}, k_{:t-1}, G)
% \end{aligned}
% \end{equation}
% % Old version
% {\bf Skills:} An option/skill $k$ is described by a tuple $(L, \pi, \beta)$, where $L$ denotes its initial condition, $\pi$ is its closed-loop control policy, and $\beta$ denotes the termination condition. An option is invoked when the state is in $L$, and the option policy $\pi$ will be executed until the termination condition is met, i.e., $\beta(\cdot)=1$. In this paper, we consider options that can be initiated in the entire state space.
% In such a mixture model, denote $\phi = \{\beta_t, k_t\}_{t=1}^T$, 
% We thus have $\pi(a_t\mid o_{:t}, k_t, G)$ as the skill (and goal) conditioned policy, and $p(\beta_t, k_t\mid \cdot)$ as the high-level policy over the skills. The posterior distribution $p(\phi \mid  \tau)$ is the distribution over skills given the trajectories. Correctly approximating that posterior will be critical to extract the skills which we will be able to reuse for new tasks.

\section{Language-guided Skill Learning with Temporal Variational Inference}
\label{sec:method}

We shall now describe our method \textbf{LA}nguage-guided \textbf{S}kill Learning with \textbf{T}emporal Variational Inference ({\bf \ours}), which jointly learns the mixture model $p(\tau) = \sum_\phi p(\phi) p(\tau \mid  \phi)$ and approximates the posterior $p(\phi \mid  \tau)$. This method is split into three steps which are shown in Fig.~\ref{fig:general}. First, given a dataset of expert demonstrations collected from multiple tasks with different goals, we use an LLM to generate an initial segmentation for these trajectories (\S\ref{sec:llm}). Then, we propose a temporal variational inference framework that aims to improve the segmentation and merge different pieces into skills (\S\ref{sec:tvi}). We also present an auxiliary training objective following the Minimum Description Length principle that helps compress the learned skills (\S\ref{sec:mdl}). Finally we introduce how the learned skills can be used to quickly solve new tasks (\S\ref{sec:onlinerl}).

\begin{figure*}[ht!]
\centering
\includegraphics[width=0.8\textwidth]
{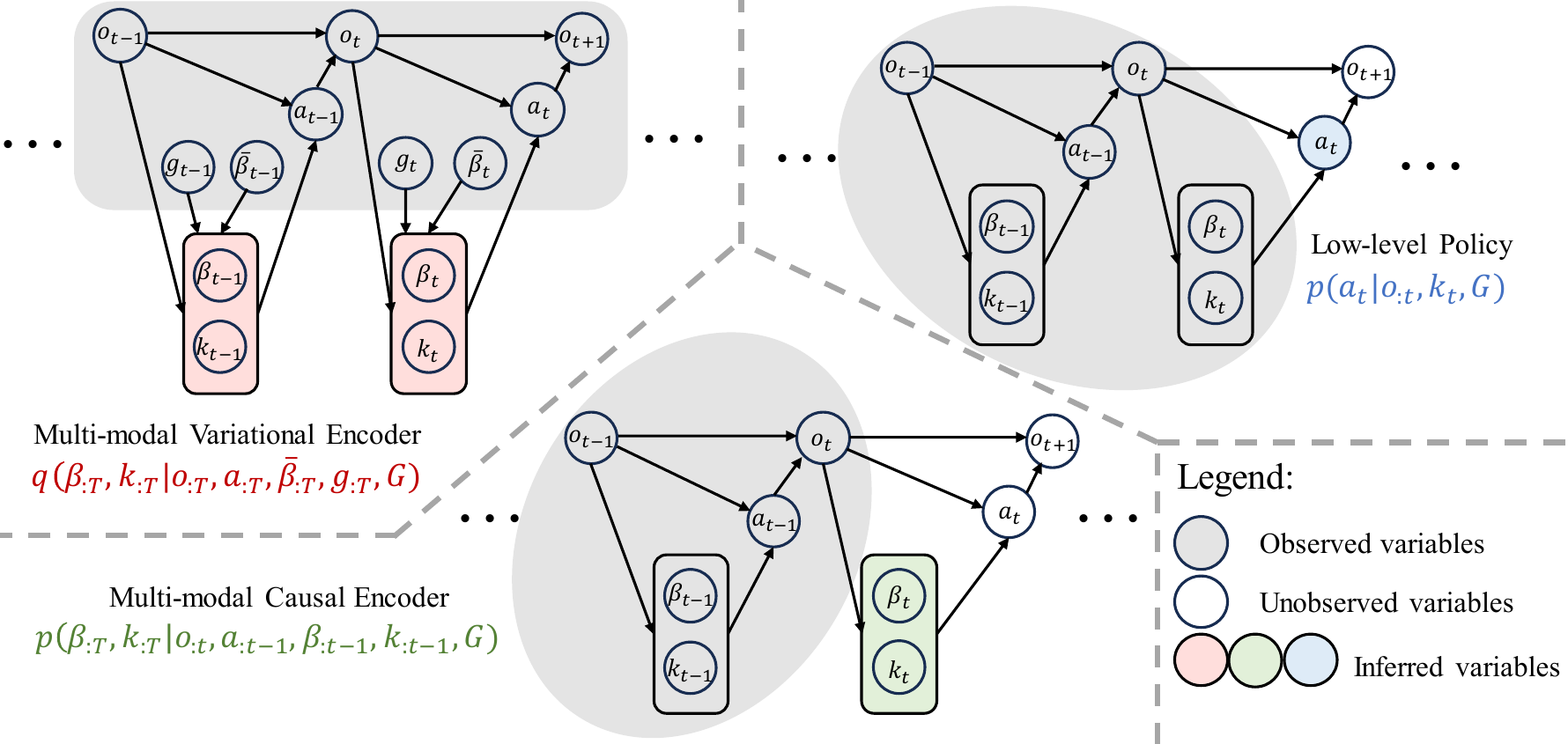}
\vspace{-0.8em}
\caption{
An overview of the probabilistic graphical model underlying \ours. Distributions are labeled with the same colors in Eqn.~\ref{eqa:obj1}. We use $q(\beta_{:T},k_{:T}\mid \cdot)$ as the approximate posterior which has access to all the information we have. $p(\beta_{:T},k_{:T}\mid \cdot)$ is the true high-level policy that is trained to mimic $q(\beta_{:T},k_{:T}\mid \cdot)$. $p(a_t\mid o_{:t}, k_t, G)$ denotes the skill-goal conditioned policy.}
\vspace{-1em}
%\caption{An overview of our Temporal Variational Inference model architecture. Each transformer corresponds to one distribution with the same color in Eqn.~\ref{eqa:obj1}.} 
\label{fig:graphical_model}
\end{figure*}

\subsection{Initial Segmentation with LLMs}
\label{sec:llm}
The core of \ours is to segment trajectories of $T$ actions into $N$ subsequences. One could imagine directly learning the graphical model. However, there are an exponential number of ways to segment a sequence of $T$ actions into $N$ subsequences, which will make the learning difficult and prone to poor local optima. Besides, the posterior $p(\phi \mid  \tau)$ will be difficult to approximate. To see this, note that, given a trajectory $\tau$, the variables $\beta_t$ are heavily correlated since switching skills at timestep $t$ affects the probability of switching skills at nearby timesteps. Hence, approximating the posterior with, e.g., a factorial distribution over the $\beta_t$ will be very inaccurate and so will be our skill extraction procedure.

To address this, we shall thus first perform a segmentation of each trajectory into subsequences by using an LLM. There is a tradeoff to achieve. On the one hand, having short subsequences will only mildly simplify the posterior distribution. On the other hand, having few, long subsequences will make the posterior easier to approximate but at the expense of a loss of capacity in our generative model, especially if the segmentation is inaccurate. As we wish to make few assumptions about the quality of the underlying LLM, we only use it to find {\bf short segments, i.e., 1 to 5 actions each}. This constraint is introduced by modifying the prompt to the LLM, as shown in Appendix~\ref{prompt example}: in the prompt, we tell the LLM that the number of actions assigned to each skill should not exceed 5 but should be larger than 1.

%old version
% As we introduced in \S~\ref{intro}, a core problem that hinders the current skill discovery methods to be applied in more complex problems is that the search space for trajectory segmentation is often too large. Given a dataset $\{\tau_i\}_{i=1}^{M}$, for each trajectory $\tau_i$, we extend the definition of termination condition $\beta$: if at timestep $t$ the agent decides that the previous skill segment should end (and thus a new skill segment should start), we set $\beta_t=1$. If the agent decides that the action taken at $t$ should be included in the previous skill segment, we set $\beta_t=0$. Consequently, a trajectory $\tau_i$ without segmentation can be represented as $\tau_i = \{G, o_1, a_1, \beta_1, o_2, a_2, \beta_2, \cdots\}$, where $\beta_t=1, \forall t$. In this paper, we also call $\beta_t$ the skill-switching variable. The total number of segments within one trajectory can be computed as $N:= \sum \beta_t$, where at the beginning, $N = T$--- the task horizon. By the definition of combination, we know that the number of possible segmentation choices is
% \begin{equation}
% \small
%     %\#\text{possible segmentation choices} = 
%     \sum_{n=1}^{N}\binom{n-1}{N-1}, 
% \end{equation}
% which has a factorial growth rate with respect to $N$. Therefore, a straightforward and efficient way to simplify this problem is to make $N$ smaller. Specifically, to identify some $\beta_t$ that should equal to $0$ given the context before moving on to the next step.

% We leverage LLMs to achieve this goal.

As shown in Fig.~\ref{fig:general} (step 1), at each iteration $i$, given a trajectory $\tau$, we prompt an LLM with the concatenation of the goal description $G$ and the action sequences $\{a_1, a_2, \cdots\}$, where each $a_t$ is described by language given the definition of the environment's action space. We ask the LLM to return the segmentation of $\tau$ and a language annotation describing each segment. We do not include the (visual) observations as input here since it would be expensive to use either the raw image or running an image captioning system. Our system does not require an optimal segmentation to start with.
%so not all information has to be used. 
For instance, given a task goal such as \cmd{place a microwaved apple slice on top of the black table} and a sequence of 44 actions, the LLM might segment that trajectory into 13 pieces, each of which is associated with a language annotation (e.g., \cmd{Navigate to kitchen step 1}).
We show a concrete example in App.~\ref{llm example}.
%(``summary action str").

% New version
The output of this step is a set of enriched trajectories where, at each timestep $t$, we added variables $\bar{\beta}_t$ and $g_t$. $\bar{\beta}_t$ equals 0 when $a_{t-1}$ and $a_t$ belonged to the same segment, 1 otherwise. $g_t$ denotes the language description generated by the LLM (\cmd{Navigate to kitchen step 1}) for the segment that contains $a_t$ ($g_t$ remains the same for the whole segment). The language annotations $g_t$ generated by the LLM are also an important conditioning information for our model to do the inference during the temporal variational inference phase. We call the first variable $\bar{\beta}$, not $\beta$, because it is not the final segmentation of the trajectory.

To simplify the learning of the graphical model, we shall make the assumption that a skill will never split a segment generated by the LLM, i.e. there will only be a merging of these initial segments into larger ones. In other words, we have $(\bar{\beta}_t = 0) \Longrightarrow (\beta_t = 0)$.
Therefore, $\sum_t \bar{\beta}_t \geq \sum_t \beta_t$.

\subsection{Skill Discovery with Temporal Variational Inference}
\label{sec:tvi}
Our goal is now to improve the initial segmentation obtained by the LLM, and merge the resulting short subsequences into longer ones that can be reused to solve new tasks efficiently, which corresponds to step 2 in Fig.~\ref{fig:general}. More specifically, we wish to jointly learn:\\
$\bullet$ $\pi(a_t\mid o_{:t}, k_t, G)$ for each possible value of $k_t$ and $G$, our goal and skill dependent policies;\\
$\bullet$ $p(\phi \mid  \tau)$, our posterior distribution over the skills and skill-switching variable, which will allow us to do skill extraction. Note that as we add the generated intermediate signals ($\bar{\beta}_t$ and $g_t$) from the first step to the trajectory $\tau$, the distribution can be further denoted as $p(\beta_{:T}, k_{:T}\mid  o_{:T}, a_{:T}, \bar{\beta}_{:T}, g_{:T}, G)$.
% \haotian{NLR: here, I'd like to be more precise with the indices over the timesteps, especially as we'll distinguish between the causal and the full model.}

% \begin{itemize}[left=0pt,itemsep=1pt,topsep=1pt]
%     \item $\pi(a_t\mid o_{:t}, k_t, G)$ for each possible value of $k_t$ and $G$, our goal and skill dependent policies
%     \item $p(\phi \mid  \tau)$, our posterior distribution over the skills and skill-switching variable, which will allow us to do skill extraction. Note that as we add the generated intermediate signals ($\bar{\beta}_t$ and $g_t$) from the first step to the trajectory $\tau$, the distribution can be further denoted as $p(\beta_{:T}, k_{:T}\mid  o_{:T}, a_{:T}, \bar{\beta}_{:T}, g_{:T}, G)$.
% % \haotian{NLR: here, I'd like to be more precise with the indices over the timesteps, especially as we'll distinguish between the causal and the full model.}
% \end{itemize}

% New version
Since we know that $\beta_t = 0$ every time $\bar{\beta}_t = 0$, we only have to compute the posterior for the timesteps $t$ such that $\bar{\beta}_t = 1$. Further, because these timesteps are less likely to be consecutive, we can hope that the posterior will be easier to approximate within our function class.

We will learn both our mixture model and approximate posterior using temporal variational inference~\citep{DBLP:journals/corr/KingmaW13}. We show the probabilistic graphical model underlying \ours in Fig.~\ref{fig:graphical_model}. We start by lower bounding the probability of a given trajectory by:
\begin{equation}
\label{eqa:elbo}
\small
\begin{aligned}
    \log p(\tau) &\geq \sum_\phi q(\phi\mid \tau)\log \frac{p(\tau, \phi)}{q(\phi\mid \tau)} \; ,
    % &= \mathbb{E}_{q(m_{:T}, g_{:T}\mid s_{:T}, a_{:T}, G)}[\sum_{t}^{T}\{\log p(m_t\mid s_{:t}, a_{:t-1}, m_{:t-1}, g_{:t-1}, G) + \log p(g_t\mid s_{:t}, a_{:t-1}, m_{:t}, g_{:t-1}, G) \\  + \log & p(a_t\mid s_{:t}, a_{:t-1}, g_t) \} - \log q(m_{:T}, g_{:T}\mid s_{:T}, a_{:T}, G)]\\
    % &=\mathbb{E}_{q(m^{\text{new}}_{:T}, k^{\text{new}}_{:T}\mid s_{:T}, a_{:T}, m_{:T}, g_{:T}, G)}\sum_{t=1}^{T}\log p(a_t\mid s_{:t}, k_{t}, G) \\ &- \sum_{t=1}^{T}\Big\{KL[q(m^{\text{new}}_t\mid s_{:T}, a_{:T}, m_t, g_{:T}, G)\\mid  p(m^{\text{new}}_t\mid s_{:t}, a_{:t-1}, m^{\text{new}}_{:t-1}, k^{\text{new}}_{:t-1}, G)] \\ &+ KL[q(k^{\text{new}}_t\mid s_{:T}, a_{:T}, m_t^{\text{new}}, g_{:T}, G)\\mid  p(k^{\text{new}}_t\mid s_{:t}, a_{:t-1}, k^{\text{new}}_{:t-1}, m^{\text{new}}_{:t}, G)] \Big\}
\end{aligned}
\end{equation}
with $q$ our approximate posterior. We will use $q$ for a factorial distribution, both over timesteps and over $\beta$ and $k$.
We factor the variational distribution as:
\begin{equation}
\label{eqa:q}
\small
\begin{aligned}
    &q(\beta_{:T}, k_{:T}\mid o_{:T}, a_{:T}, \bar{\beta}_{:T}, g_{:T}, G) = \\
    &\textstyle\prod_{t=1}^{T} q(\beta_t\mid o_{:T}, a_{:T}, \bar{\beta}_t, g_{:T}, G)q(k_t\mid o_{:T}, a_{:T}, \beta_t, k_{t-1}, g_{:T}, G),
\end{aligned}
\end{equation}
% further enforcing $k_t=k_{t-1}$ when $\beta_t=0$ to match the generative model. 
% Note that both $q(k_t\mid \cdot)$ and $q(\beta_t^{\text{new}}\mid \cdot)$ are conditioned the whole trajectory (history + future)
Due to space limit, we show the factorized form of $p(\tau, \phi)$ in App.~\ref{app:tvi}, Eqn.~\ref{eqa:p}. Substituting the corresponding terms in Eqn.~\ref{eqa:elbo} with Eqn.~\ref{eqa:p} and Eqn.~\ref{eqa:q}, we get the following objective:
\begin{equation}
\label{eqa:obj1}
\small
    \begin{aligned}
        J(\theta) \approx& \mathbb{E}_{q_{\theta}(\phi\mid \tau)}\sum_{t=1}^{T}\log \textcolor{darkblue}{\pi_{\theta}(a_t\mid o_{:t}, k_{t}, G)} \\
        - \sum_{t=1}^{T}\Big\{&\text{KL}[\textcolor{darkred}{q_{\theta}(\beta_t\mid o_{:T}, a_{:T}, \bar{\beta}_t, g_{:T}, G)} \| \\ 
         &\quad\:\:\:\textcolor{darkgreen}{p_{\theta}(\beta_t\mid o_{:t}, a_{:t-1}, \beta_{:t-1}, k_{:t-1}, G)}] \\ 
        + &\text{KL}[\textcolor{darkred}{q_{\theta}(k_t\mid o_{:T}, a_{:T}, \beta_t, k_{t-1}, g_{:T}, G)}\| \\ 
         &\quad\:\:\:\textcolor{darkgreen}{p_{\theta}(k_t\mid o_{:t}, a_{:t-1}, k_{:t-1}, \beta_{:t}, G)}] \Big\},
    \end{aligned}
\end{equation}
where we parameterize all models with $\theta$. As we will introduce below, we use the same transformer to model the distributions highlighted with the same color. Note that while both $p_{\theta}$ and $q_{\theta}$ infer the distribution over $\beta_t$ and $k_t$; $q_{\theta}$ is conditioned on the whole trajectory (history and future) and the generated language annotations; $p_{\theta}$ is only conditioned on the history. We add such causal constraints to ensure the inference policy can be used online where only the information till the current step is accessible.

Intuitively, we want $q_{\theta}$ to perform the best inference of the skills by conditioning on all the information we have. Meanwhile, we want $p_{\theta}$ , the true high-level policy to be used on new tasks, to mimic the results of $q_{\theta}$ without the generated language annotations and future information, because both of which are absent during online testing. The first term of $J(\theta)$ encourages $q_{\theta}$ to generate proper skill-switching variables $\beta_t$ and skills $k_t$ such that we can train a policy $\pi$ to accurately predict the actions conditioned on these latent variables. The other two KL-divergence terms serve two purposes. First, they encourage the inference $p_{\theta}$ of $\beta_t$ and $k_t$ given only the history to be as close as possible to the inference $q_{\theta}$ given all the information; second, they serve as regularization terms for $q_{\theta}$ to avoid overfitting.

After this temporal variational inference training stage, we have obtained a low-level policy $\pi_{\theta}$ that predicts actions given the observations and a specific skill, and a high-level policy $p_{\theta}$ that predicts the skill given the information available at the current timestep. 

{\bf Constraints on inferring $\beta_t$}. We made the assumption that no further segmentation is needed after step 1 and the agent only needs to merge these segments. To ensure this is met in practice,
% i.e. we let the agent reconsider whether it should infer a new skill ($\beta_t=1$) only at those timesteps $t$ where the results from step 1 (LLM) say that it should switch to a new skill ($\bar{\beta}_t=1$). 
we sample $\beta_t$ and $k_t$ with the following process in step 2:
\begin{equation}
\small
\label{eqa:assump}
    \begin{aligned}
        % &\beta^{\text{new}}_t \sim \beta_t q_{\theta}(\beta^{\text{new}}_t\mid o_{:}, a_{:}, \beta_t, g_{:}, G)\\
        % &k_t \sim \beta^{\text{new}}_t q_{\theta}(k_t\mid o_{:}, a_{:}, \beta_t^{\text{new}}, g_{:}, G) + (1-\beta^{\text{new}}_t) k_{t-1}
        &\beta_t \sim \bar{\beta}_t q_{\theta}(\beta_t\mid o_{:T}, a_{:T}, \bar{\beta}_t, g_{:T}, G),\\
        &k_t \sim \beta_t q_{\theta}(k_t\mid o_{:T}, a_{:T}, \beta_t, g_{:T}, G) + (1-\beta_t) \delta(k_t==k_{t-1}),
    \end{aligned}
\end{equation}
where $\delta$ denotes the delta function. The same strategy applies to $p_{\theta}$ as well. This greatly simplifies the inference problem and enables the agent to efficiently discover the reusable skills. However, it also requires the segmentations made in step 1 to be maximally detailed, so that we do not need to further break down the subsequences in step 2.

\subsection{Minimum Description Length for Skills}
\label{sec:mdl}
%As mentioned above, 
While jointly learning the posterior distribution and skill-conditioned policies, we would like to merge the resulting subsequences from step 1 into longer and reusable ones. That is, we would like to further compress the results and decrease the number of subsequences of each trajectory. Inspired by~\citet{DBLP:conf/nips/JiangLEKF22}, we introduce an auxiliary compression objective, following the Minimum Description Length (MDL) Principle~\citep{DBLP:journals/automatica/Rissanen78}. 

MDL favors the simplest model that accurately fits the given data, which provides guidance for finding the common structures inside the data and further compress them. In this context, MDL suggests choosing the skills that provides the shortest description of the trajectories. We consider the two-part form of MDL:
\begin{equation}
\small
    \phi^* = \text{argmin}_{\phi} L(\mathcal{D}\mid \phi) + L(\phi),
\end{equation}
where $L(\mathcal{D}\mid \phi)$ denotes the number of bits required to encode the trajectories $\mathcal{D}$ given the model $\phi$, and $L(\phi)$ is the number of bits required to encode the model $\phi$ itself. Computing $L(\phi)$ generally requires approximating the complexity of neural networks. In this paper, we simply add one constraint to the training process as an approximation: we set a maximum number of skills for the skill library so that the number of skills used to encode all the trajectories cannot exceed this number. After training we only keep the skills that are used frequently enough (over some threshold) and use them for online testing.

We focus on minimizing the first term $L(\mathcal{D}\mid \phi)$. Recall that we use $\phi = \{\beta_t, k_t\}_{t=1}^{T}$ to encode the trajectory $\tau = \{G, o_t, a_t, \bar{\beta}_t, g_t\}_{t=1}^{T}$. According to optimal code length theory~\citep{DBLP:books/wi/01/CT2001}, the expected number of bits of code generated by $p(k)$ is: $\mathbb{E}_k[-\log p(k)]$. Applying the formulation defined in \S~\ref{sec:tvi}, we can derive the following objective (see Appendix~\ref{app:abl} for detailed derivations):
\begin{equation}
\small
\begin{aligned}
   &\mathcal{L}_{\text{MDL}}(\theta) = -\sum_t \mathbb{E}_k \log [q_t(k\mid \cdot)q_t(\beta_t=1\mid \cdot) \\& 
   \quad\quad\quad\quad\quad\quad\quad\quad\quad\quad\:+ q_t(\beta_t=0\mid \cdot)\mathbbm{1}(k==k_{t-1})],
    \end{aligned}
\end{equation}
where $q_t(k\mid \cdot)$ and $q_t(\beta_t\mid \cdot)$ refer to the variational posterior (highlighted in red in Eqn.~\ref{eqa:obj1}). Intuitively, by minimizing this objective, the agent is encouraged to infer fewer skills for each trajectory (increase $\sum_t q_t(\beta_t=0\mid \cdot)$), as well as to increase the average confidence for choosing the skill (decrease $-\mathbb{E}_k \log q_t(k\mid \cdot)$) when it has to. A discussion of the difference between the proposed objective and the objective proposed in~\citet{DBLP:conf/nips/JiangLEKF22} can be found in App.~\ref{app:mdl}. In general, 
our objective function is able to more accurately reflect the skill switches's influence on the code length and also increase trainability, since our formulation and derivation do not ignore the effect of the skill $k_{t-1}$ chosen at last time step $t-1$.

The overall training objective for step 2 is:
\begin{equation}
\label{eqa:jointobj}
\small
    \max_{\theta} J(\theta) - \lambda L_{\text{MDL}}(\theta),
\end{equation}
where $\lambda$ is a hyperparameter that controls the weight of the auxiliary compression objective.

\subsection{Practical Implementation and Model Architecture}

We show the model architectures that implement the three distributions in App. Fig.~\ref{fig:architecture}.
As an overview, we use three transformers to parameterize the three distributions in Eqn.~\ref{eqa:obj1} (in three colors) respectively. The \textcolor{darkred}{Multi-modal Variational Encoder} approximates $q(\beta, k\mid  \cdot)$, which has access to all information in each trajectory. The \textcolor{darkgreen}{Multi-modal Causal Encoder} approximates $p(\beta, k\mid  \cdot)$, which has access to information only up to the current step (implemented using a causal transformer). %Thus we use a causal transformer to model it. 
The \textcolor{darkblue}{Low-level Policy} is also a causal transformer that approximates $\pi(a\mid  o, k, G)$. We use individual pretrained encoders for the multi-modal input, and then train a linear encoder for each modality to map the inputs into the same embedding space.

% \begin{figure}[htbp]
% \centering
% \includegraphics[width=0.9\linewidth]{icml2024networks.pdf}
%     \caption{An overview of our Temporal Variational Inference Model Architecture. Each transformer corresponds to one distribution with the same color in Eqn.~\ref{eqa:obj1}.} 
%     \label{fig:graphical_model}
% \end{figure}

We model the skills $k$ as discrete variables and set the maximum number of skills to be $K$. 
% It is less straightforward to compute the MDL loss for continuous variables so we leave this for future work. 
%To update the joint objective in Eqn.~\ref{eqa:jointobj}, in practice, at the beginning of training we have a warm-up stage.
To stabilize the optimization with the joint objective in Eqn.~\ref{eqa:jointobj}, we perform a warm-up stage at the beginning of training.
Specifically, for a certain amount of episodes, we set $\lambda=0$ which stops the gradients from the MDL loss to $q_{\theta}$ in Eqn.~\ref{eqa:obj1}. We have this pretraining stage to prevent uninformative gradients from $q$ at the beginning of training, and also prevent from the model quickly converge to inferring the whole trajectory as a single skill because of the gradients from the MDL term.

\subsection{Online Hierarchical RL}
\label{sec:onlinerl}
Now we introduce how we use the learned skills for online hierarchical RL. Following prior skill learning approaches, we {\bf freeze} the low-level policy $\pi(a\mid  \cdot)$ and train a high-level control policy that directly outputs skills $k$. Changing the action space from the primitive action space to the learned skill space yields efficient adaptation as the skills are temporally-extended, which can both shorten the task horizon and perform structural exploration. 

Specifically, we let the agent learn a high-level control policy $\pi_{\psi}(k_t\mid  o_{:t}, a_{:t-1}, k_{:t-1}, \beta_{:t}, G)$ and termination condition $\pi_{\psi}(\beta_t\mid  o_{:t}, a_{:t-1}, k_{:t-1}, \beta_{:t-1}, G)$ by fine-tuning on $p(k_t\mid  \cdot)$ and $p(\beta_t\mid  \cdot)$. We train the policy by maximizing the following objective:
\begin{equation}
\small
    \begin{aligned}
        &\mathcal{J}(\psi) = \mathbb{E}_{\pi_{\psi}}[\sum_t \gamma^t (r_t \\&- \alpha_1 \text{KL}(\pi_{\psi}(k_t\mid o_{:t}, a_{:t-1}, k_{:t-1}, \beta_{:t}, G)\|p(k_t\mid  \cdot)) \\&-\alpha_2 \text{KL}(\pi_{\psi}(\beta_t\mid o_{:t}, a_{:t-1}, k_{:t-1}, \beta_{:t-1}, G)\| p(\beta_t\mid \cdot)))],
    \end{aligned}
\end{equation}
where we augment the standard maximizing cumulative return objective with additional KL regularization between the high-level policy and the offline pretrained distributions over the skills and termination conditions. This is to ensure the high-level policy remains close to the pre-learned prior over the skills. We adopt Soft Actor-Critic (SAC)~\citep{DBLP:conf/icml/HaarnojaZAL18} with Gumbel-Softmax~\citep{DBLP:conf/iclr/JangGP17} (for the categorical action distribution) to train the agent to maximize the cumulative return. During sampling, at the first step of each episode, the agent will first choose a skill $k_1$ according to $\pi_{\psi}(k_1\mid \cdot)$, using the skill to interact with the environment through $\pi_{\theta}(a\mid \cdot)$ until the termination condition $\beta_t=1$ is met, and then switches to the next skill.

\section{Experiments}
Our experiments focus on, 1. Can \ours discover distinct and semantically meaningful skills from a set of demonstrations? 2. How well do the hierarchical policies transfer to new tasks (zero-shot)? 3. Do the learned skills help accelerate learning on downstream tasks? 4. How do components of \ours contribute to its overall performance? 
% \footnote{We show demo videos at \href{https://language-skill-discovery.github.io/}{language-skill-discovery.github.io}.}

\subsection{Experimental Setup}

\begin{figure*}[htbp]
\centering
\includegraphics[width=0.82\textwidth]
{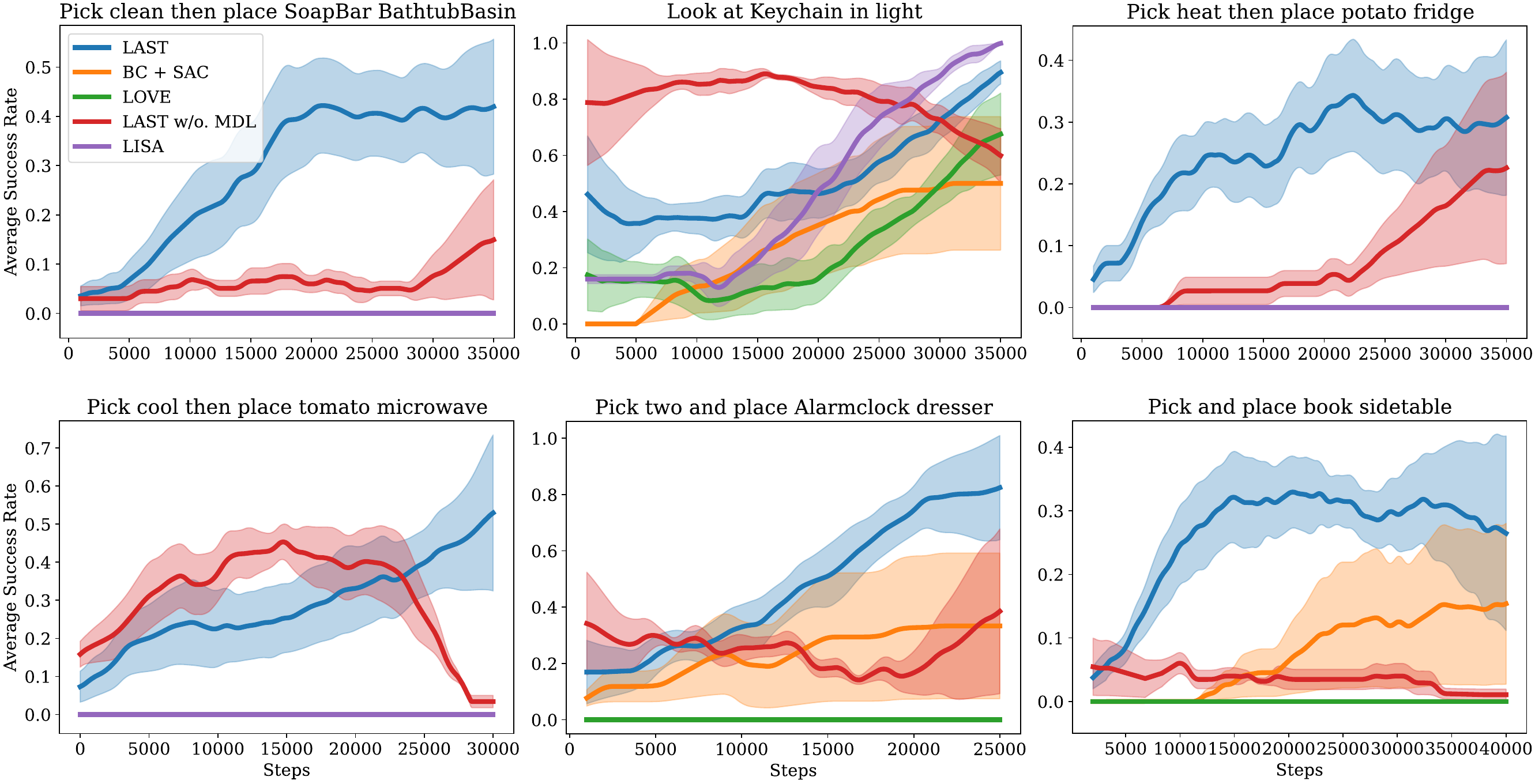}
\vspace{-0.8em}
\caption{Comparison results of our method \ours against other baselines in six downstream tasks of ALFRED. We plot average success rate v.s. timesteps with 95\% confidence interval error bar ($\geq 5$ seeds).} 
\vspace{-0.8em}
\label{fig:rlfinetune}
\end{figure*}

{\bf BabyAI}~\citep{DBLP:conf/iclr/Chevalier-Boisvert19} is an environment where an agent navigates and interacts in a grid world to achieve a goal described in language (e.g., \cmd{open a red door and go to the ball on your left}). For our experiments, we use the environment's symbolic partially observable state space, where the state vector describes the type and color of each grid cell. At each time step, the agent can choose from 6 actions including pickup/drop/toggle an object. The BabyAI dataset contains expert demonstrations collected from 40 different task types of varying levels of difficulty. We uniformly at random sample 100 trajectories from each task type perform \ours. Note that compared to previous imitation learning work on BabyAI, our setting is much harder in a few key ways, 1) We sample a very small subset of trajectories to train our model unlike previous work often requiring up to 1M trajectories. 2) We use \ours on data from multiple tasks jointly. This is in contrast to prior work, where each task type requires learning its own separate policy. 

{\bf ALFRED}~\citep{DBLP:conf/cvpr/ShridharTGBHMZF20} is a complex environment based on the AI2-THOR~\citep{DBLP:journals/corr/abs-1712-05474} simulator and is a domain where an agent is required to perform diverse household tasks following a goal described by natural language. The environment itself contains $100+$ floorplans and $100+$ objects to interact with. The observation space consists of the agent's egocentric view (image) of the current environment. The action space consists of 12 discrete action types (e.g., \cmd{move forward}, \cmd{turn right}, \cmd{toggle}) and 82 discrete object types. At each time step the agent needs to choose an action type along with an object type to interact with the environment. For the ALFRED dataset, we follow the settings in~\citet{DBLP:conf/iccv/PashevichS021} where the training dataset consists of more than 20000 trajectories. Each trajectory includes a goal description and a sequence of observations and actions. Note that different from many previous work on ALFRED (especially the leaderboard methods) {\bf we do not use the step-by-step instructions provided by environment (neither during training nor testing)}, which gives us a standard goal-conditioned RL setting and is potentially more practical as instructions annotated by humans could be expensive.
We want to highlight that the goal of this work is not to directly compare to systems on the ALFRED leaderboard in terms of overall performance, rather, we adopt ALFRED as a challenging testbed demonstrating our proposed skill discovery method. 

We use GPT-4 as the LLM to generate the initial segmentation in all our experiments\footnote{As an alternative, we have used open-sourced LMs (e.g., Phi-2 and Mistral 7B) to generate the initial segmentation. However, 1) they support insufficient context length and 2) they often fail to follow instructions on generating output with specific format. }. In ALFRED, we use pretrained T5 encoder~\citep{DBLP:journals/jmlr/RaffelSRLNMZLL20} and Faster R-CNN encoder~\citep{DBLP:conf/nips/RenHGS15} to preprocess the image and language data.
Once again, we do not aim to compare various LLMs and encoders, we adopt the strongest systems off-the-shelf and focus on our system described in \S~\ref{sec:method}, which is presumably orthogonal from specific model choice. 

\begin{table}[t!]
\footnotesize
%\scriptsize
\centering
%\vspace{-0.8em}
\caption{Online (zero-shot) task success rate comparison on BabyAI (\textbf{top}) and ALFRED (\textbf{bottom}). Note that the subtasks in ALFRED have short horizons (i.e., subgoals of each individual task which requires significantly fewer steps to reach). In comparison, in the downstream task learning setting, the agent is tested on six individual tasks with the full length.}
\begin{tabular}{lllllll}
\label{tab:babyai_alfred_zero}
\centering
Model & \rotatebox{90}{BossLevel} & \rotatebox{90}{MiniBoss} & \rotatebox{90}{Synth} & \rotatebox{90}{SynthLoc} & \rotatebox{90}{SynthSeq} & \rotatebox{90}{Average}\\\hline 
 % Ours (valid unseen) & 52 & 82 & 65 & 82 & 63 & 53 & 48 & 37 & 44 \\
 BC & 25 & 23 & {\bf 53} & 48 & 20 & 34  \\
 LOVE & 27 & 31 & 40 & 63 & 25 & 37  \\
 %Ours w/o. MDL & 38 & 38 & 42 & 54 & 28 & 40 \\
 LISA & 27 & 26 & 49 & 49 & 27 & 36  \\
  {\bf \ours} & {\bf 36} & {\bf 34} & 48 & {\bf 66} & {\bf 32} & {\bf 43} \\\hline
\end{tabular}
%\vspace{3em}
\scriptsize
\begin{tabular}{llllllllll}
\centering
Model & \rotatebox{90}{Avg} & \rotatebox{90}{Clean} & \rotatebox{90}{Cool} & \rotatebox{90}{Heat} & \rotatebox{90}{Pick} & \rotatebox{90}{Put} & \rotatebox{90}{Slice} & \rotatebox{90}{Toggle} & \rotatebox{90}{GoTo}  \\\hline 
 % Ours (valid unseen) & 52 & 82 & 65 & 82 & 63 & 53 & 48 & 37 & 44 \\
 BC & 45 & 66 & 31 & 36 & 57 & 47 & 38 & 32 & 41 \\
 $(SL)^3$ & 34 & 68 & 82 &75 & 50 & 45 & {\bf 55} & 32 &15 \\
 seq2seq & 17 & 16 & 33 & 64 & 20 & 15 & 25 & 13 & 14 \\
  seq2seq2seq & 26 & 15 & 69 & 58 & 29 & 42 & 50 & 32 & 15 \\
  seq2seq & 17 & 16 & 33 & 64 & 20 & 15 & 25 & 13 & 14 \\
  LOVE & 47 & 76 & {\bf 92} & 41 & 58 & 57 & 51 & 28 & 38 \\
  LISA & 46 & 73 & 82 & 21 & {\bf 66} & 49 & 35 & 27 & 38 \\
   {\bf \ours} & {\bf 52} & {\bf 82} & 65 & {\bf 82} & 63 & {\bf 53} & 48 & {\bf 37} & {\bf 44} \\\hline
\end{tabular}

\end{table}

\subsection{Zero-shot Transfer}
\label{exp:zeroshot}
We first study the zero-shot transfer performance on BabyAI and ALFRED. 
For BabyAI, we compare with standard Behavior Cloning (BC) augmented with some inductive bias for object predictions as we explained in appendix~\ref{app:implementation}; 
LOVE~\citep{DBLP:conf/nips/JiangLEKF22}, the state-of-the-art method for learning skills from demonstrations (without language) which first proposes the compression objective; and 
LISA~\citep{DBLP:conf/nips/GargVKSE22}, a language-based skill learning method.
We further equip LISA with our network architectures and provide it with the initial segmentation results. It can thus be seen as an simpler version of our algorithm without temporal variational inference and directly mapping the initial segmentation results to skills. We do not compare to~\citep{DBLP:conf/icml/Shankar020} since it is similar to LOVE but without the compression term. We implement all the baselines using the same transformer structure we use for \ours to make fair comparison. 
% For ALFRED, since we are using a setting where step-by-step human-annotated instructions are not given, the best method we are able to find in a similar setting is $(SL)^3$~\citep{DBLP:conf/acl/Sharma0A22}. 
For ALFRED, since our setting does not assume the access to the step-by-step human-annotated instructions, the best performing method we are able to find in a similar setting is $(SL)^3$~\citep{DBLP:conf/acl/Sharma0A22}. 
We also compare against seq2seq and seq2seq2seq, both of which are methods provided by the original paper of ALFRED in the same setting~\citep{DBLP:conf/cvpr/ShridharTGBHMZF20}, as well as BC (modified using the same policy network architecture of \ours).

We show the results in Table~\ref{tab:babyai_alfred_zero} (top: BabyAI, bottom: ALFRED). On both domains, \ours can achieve the highest average success rate and outperforms the baselines in most tasks, indicating that the skills accompanied with the hierarchical policies our method learns can be well transferred on diverse new tasks. We notice that the largest performance gain of \ours in ALFRED comes from the "GoTo" tasks, which is the navigation tasks that previous language-based methods find hard to solve~\citep{DBLP:conf/acl/Sharma0A22}.

\begin{figure*}[t!]
\centering
    \includegraphics[width=0.5\linewidth]{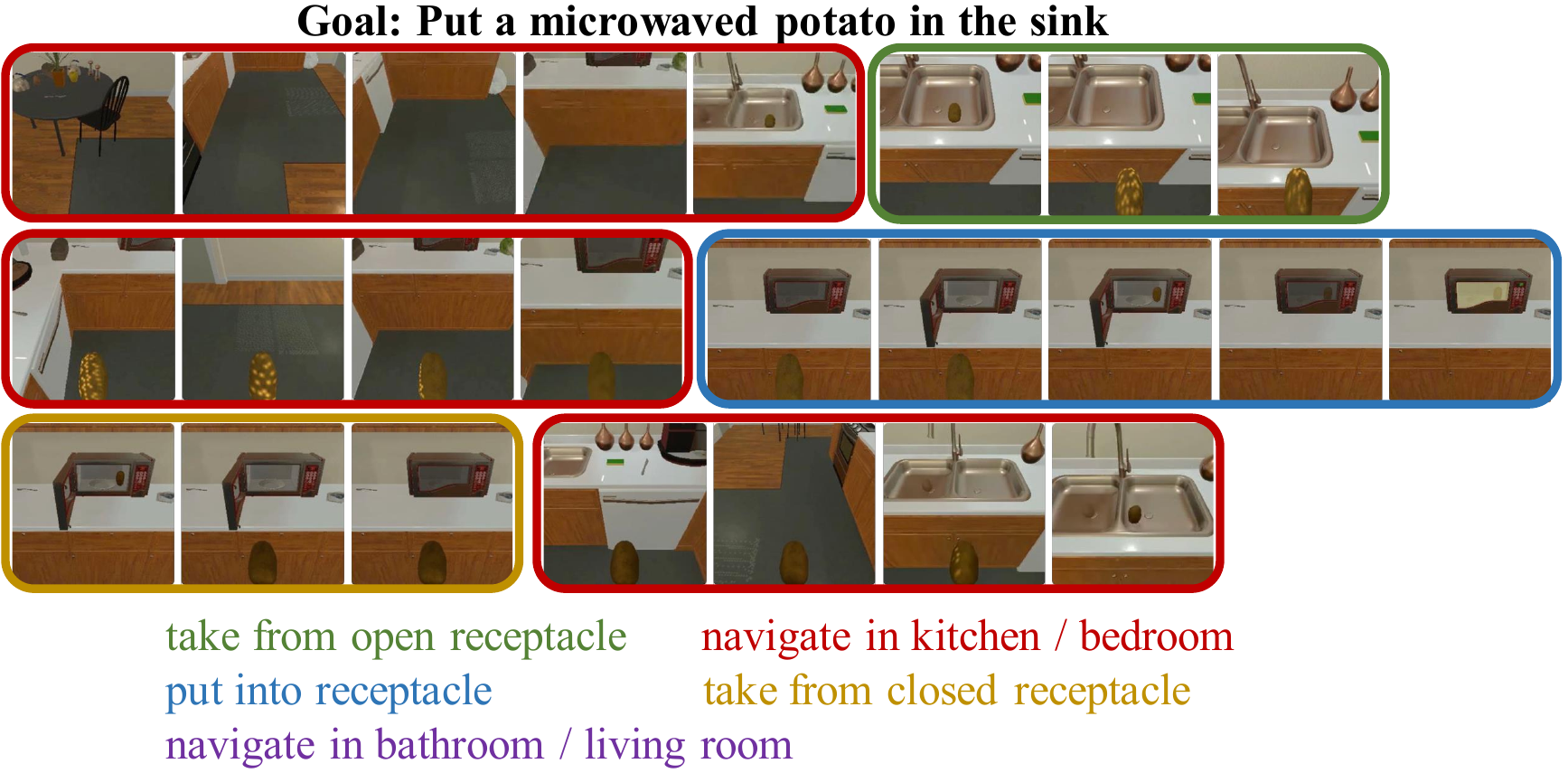}
    \includegraphics[width=0.48\linewidth]{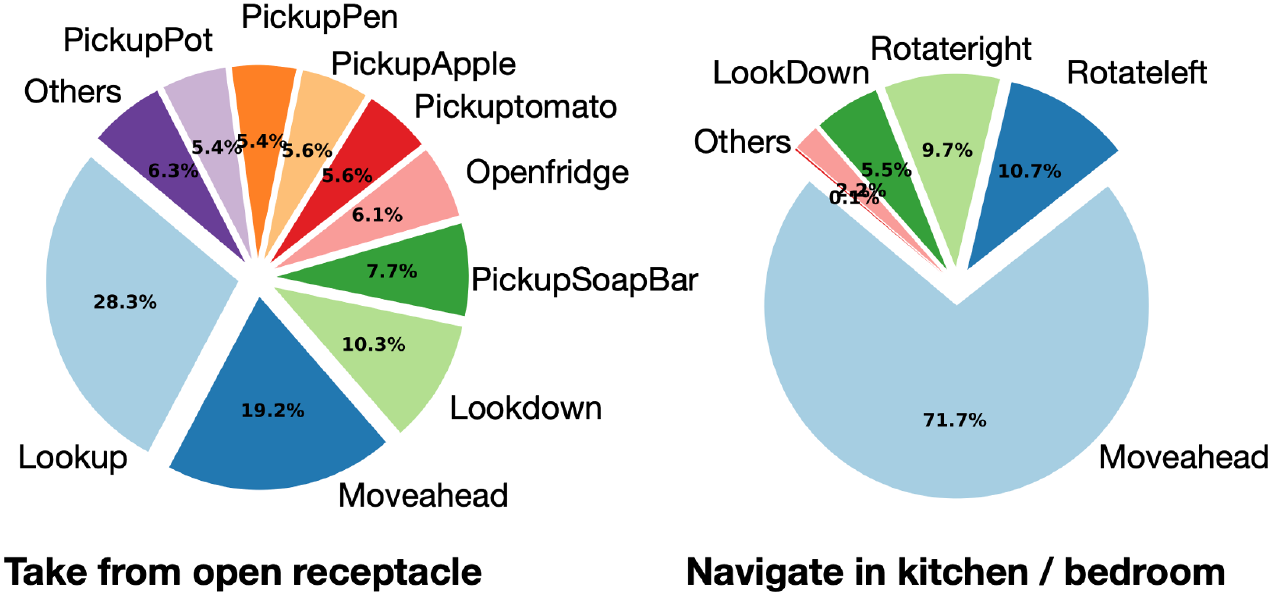}
    \vspace{-0.5em}
    \caption{\textbf{Left}: \ours's skill segmentation for task \cmd{Put a microwaved potato in the sink}. 
    %We also show another example subsequence from different tasks for each used skill. 
    %We provide more example trajectories and their segmentation in App. Fig.~\ref{fig:appqualt}. 
    \textbf{Right}: Example of discovered skills and their most-commonly used actions.
    %We show details of all five common skills in App. Fig.~\ref{fig:app_pie}.} 
    We show more trajectories, segmentations, and common skills in App. Fig. \ref{fig:appqualt} and \ref{fig:app_pie}.} 
    \vspace{-0.5em}
    \label{fig:qualitative}
\end{figure*}

\subsection{Learning on Downstream Tasks}
We further test whether the learned skills can facilitate the agent's learning on different downstream tasks. We randomly pick one long-horizon task from each of the six task categories in ALFRED and run online RL fine-tuning with SAC~\citep{DBLP:conf/icml/HaarnojaZAL18}. Note that learning from scratch using SAC is not able to achieve any success in these six tasks. We compare our method with the same set of methods introduced in the last section, as well as an additional baseline that does not include the MDL term as the auxiliary training loss during temporal variational inference.  

As shown in Fig.~\ref{fig:rlfinetune}, \ours outperforms baselines in 5 out of 6 tasks, demonstrating that the learned skills can facilitate downstream task learning for multiple types of tasks. In \textit{Pick clean}, \textit{Pick heat} and \textit{Pick cool} tasks, where the task horizon ($\sim 200$) is significantly larger, only \ours-based methods can achieve positive success rate. Other skill discovery baselines like LOVE and LISA perform comparably only on \textit{Look at Keychain in Light} task, which has the shortest task horizon ($<50$) among the six tasks. The results indicate that both the initial segmentation phase (compared with LOVE) and the temporal variational inference phase (compared with LISA) are important to learn skills that can solve complex tasks.
We further note that without the MDL term, \ours is still able to learn skills that can help downstream learning compared to the other baselines, however, the absence of the MDL term decreases \ours's performance.

Ablation study results can be found in App.~\ref{app:abl}, where we further investigate the influence of the initial segmentation, the generated language annotation $g$, MDL objective, replacing initial segmentation with the one provided by the dataset's human annotated instructions and the constraints on inferring $\beta_t$ (Eqn.~\ref{eqa:assump}).

\subsection{Qualitative Evaluation of the Learned Skills}

Videos of the learned skills can be found at \href{https://language-skill-discovery.github.io/}{language-skill-discovery.github.io}. We aim to gain a better understanding on the skills discovered by \ours by investigating the sub-trajectories segmented by it across three runs with different random seeds.
We notice that there are five skills being discovered in all three runs.
We manually identify the semantics of these five skills by looking at their corresponding sub-trajectories,
we list them in Fig.~\ref{fig:qualitative} (left).
%, we also provide videos of the sub-trajectories in our~\href{https://language-skill-discovery.github.io/}{website}.
%They are: 1) \textcolor{darkgreen}{take from open receptacle}; 2) \textcolor{darkred}{navigate in kitchen / bedroom}; 3) \textcolor{darkblue}{put into receptacle}; 4) \textcolor{darkyellow}{take from closed receptacle}; and 5) \textcolor{darkpurple}{navigate in bathroom / living room}.
%We provide videos of the sub-trajectories in our~\href{https://language-skill-discovery.github.io/}{website}.
To better demonstrate this finding, in Fig.~\ref{fig:qualitative} (left), we show a trajectory, segmented and labelled by \ours, with our interpretation of the skills.
In this example, the trajectory has been split into six sub-trajectories of varying length. 
In which, for example, the sequence of actions to take a potato from the sink has been labeled as the \textcolor{darkgreen}{take from open receptacle} skill, and thus it may share the skill representation with sub-trajectories that take object from a table in another trajectory.

In Fig.~\ref{fig:qualitative} (right), we statistically show the proportion of each ALFRED action being included in the segments correspond to the \textcolor{darkgreen}{take from open receptacle} and \textcolor{darkred}{navigate in kitchen / bedroom} skills. 
For example, for skill \textcolor{darkgreen}{take from open receptacle}, action \cmd{Pickup} appears frequently; while in the \textcolor{darkred}{navigate in kitchen / bedroom} skill, more than 70\% of the actions are \cmd{Moveahead}.
We are delighted to see that in addition to being helpful quantitatively (as shown in previous subsections), the skills discovered by \ours could to some extent be interpreted by humans, we believe this is due to our design that leveraging LLMs in skill/segment initialization.
We provide more example trajectory segments in App. Fig.~\ref{fig:appqualt}. {\bf As we mentioned before, in ALFRED we find there are five categories of skills that are always discovered by our approach across different random seeds and we include the pie charts for all of them in App. Fig.~\ref{fig:app_pie}.}  We also show the transition probability matrix between discovered skills in App. Fig.~\ref{fig:appd_trans}.

% Videos of the learned skills and segmented trajectories can be found in~\href{TODO}{our website}.We show an example of the skill segmentation inferred for a trajectory in Fig.~\ref{fig:qualt1}. In the same figure, we also show some segments from other trajectories that are inferred as the same skill. More examples can be found in App.~\ref{sec:appqualt}. As we also show in Fig.~\ref{fig:pie}, among different random seeds, we find that there are 5 skills that are commonly discovered by \ours, including taking out some object from an open/closed receptacle, navigation in different rooms, as well as put something in a receptacle. For the navigation skills, we can see from the pie chart that action ``Move ahead'' is used most frequently. For skill ``taking our something from an open receptacle'', action ``Pick up'' is used more frequently while for ``taking our something from an closed receptacle'', more actions for open/close a receptacle are involved. We also show a skill transition ``heat map'' in Fig.~\ref{fig:trans}. We can see that in this skill set, there are another six skills discovered besides the five most common skills but the transition probabilties from the other skills to them in general are much lower than the first five skills.

\section{Discussion}
In this paper, we studied the problem of learning skills from demonstrations. 
We propose {\ours}, a framework that learns reusable skills from expert trajectories by 1) querying LLMs for an initial segmentation; 2) leveraging temporal variational inference to merge subsequences into skills; 3) training with an auxiliary MDL term that further compresses the skills.
%Through the proposed framework, {\ours}, the search space for reusable skills from expert trajectories are gradually made smaller by: 1. utilizing LLMs to do an initial segmentation of the trajectories, 2. leveraging temporal variational inference to improve the segmentation and merge subsequences into skills, 3. training with an auxiliary MDL term that further compresses the skills.
% We also provided an approach to use these skills to accelerate online reinforcement learning on new tasks. 
We found empirically that \ours enables the agent to learn semantically meaningful skills that can help solve long-horizon complex tasks.

The proposed initial segmentation approach requires agents equipped with discrete action tokens. This is one limitation of the proposed method. However, we would like to point out that many practical applications have the same settings (e.g., policies that use API calls or tools, AI assistants that help with typical performance in a program or OS, gaming settings such as Atari where actions tend to be discretized). Besides, we should also note that the core idea of initial segmentation is to leverage the commonsense knowledge from foundation models to perform the initial segmentation. At the moment, we are limited to using Large Language Models (hence using the text description of the actions). But it is also possible to use large video foundation models to improve the initial segmentation and extend our work to other domains in the future.

% \clearpage
\section*{Acknowledgement}
This work was partially done while Haotian Fu, Pratyusha Sharma,  and Elias Stengel-Eskin were interning at MSR. The authors would like to thank Sumana Basu, Ayush Jain, Yiding Jiang, Michael Littman, Vincent Micheli, Benjamin Spiegel, Tongzhou Wang, Tian Yun, Zilai Zeng, ICML anonymous reviewers, as well as other colleagues and friends at Microsoft Research and Brown University for discussions and helpful feedback. This work was supported in part by NSF grant \#1955361, CAREER award \#1844960 to Konidaris, and ONR grant \#N00014-22-1-2592. And Nicolas Le Roux was supported by a CIFAR AI chair.
\section*{Impact Statement}
\label{section:broader_impact}
We do not foresee significant societal impact resulting from our proposed method. 
The proposed \ours system leverages LLMs to label pre-collected trajectory datasets and then learns to merging the segments into reusable skills. 
Despite \ours does not directly interact with humans in any way, caution needs to be exercised if one extends our work to a setting where humans are more involved. 
We use pre-trained LLMs for obtaining the initial segments and their labels, this could potentially bring hallucinations into the system. 
Although our Temporal Variational Inference framework is designed to refine and merge the LLM-generated segments into more reusable skills, mis-information from the initial labels could nonetheless have an effect on the converged skills.

\bibliography{example_paper}

\begin{thebibliography}{64}
\providecommand{\natexlab}[1]{#1}
\providecommand{\url}[1]{\texttt{#1}}
\expandafter\ifx\csname urlstyle\endcsname\relax
  \providecommand{\doi}[1]{doi: #1}\else
  \providecommand{\doi}{doi: \begingroup \urlstyle{rm}\Url}\fi

\bibitem[Ajay et~al.(2021)Ajay, Kumar, Agrawal, Levine, and Nachum]{DBLP:conf/iclr/AjayKALN21}
Ajay, A., Kumar, A., Agrawal, P., Levine, S., and Nachum, O.
\newblock {OPAL:} offline primitive discovery for accelerating offline reinforcement learning.
\newblock In \emph{9th International Conference on Learning Representations, {ICLR} 2021, Virtual Event, Austria, May 3-7, 2021}. OpenReview.net, 2021.

\bibitem[Anderson et~al.(2018)Anderson, Wu, Teney, Bruce, Johnson, S{\"u}nderhauf, Reid, Gould, and Hengel]{anderson}
Anderson, P., Wu, Q., Teney, D., Bruce, J., Johnson, M., S{\"u}nderhauf, N., Reid, I., Gould, S., and Hengel, A.~V.
\newblock Vision-and-language navigation: Interpreting visually-grounded navigation instructions in real environments.
\newblock \emph{2018 IEEE/CVF Conference on Computer Vision and Pattern Recognition}, pp.\  3674--3683, 2018.

\bibitem[Andreas et~al.(2017)Andreas, Klein, and Levine]{pmlr-v70-andreas17a}
Andreas, J., Klein, D., and Levine, S.
\newblock Modular multitask reinforcement learning with policy sketches.
\newblock In Precup, D. and Teh, Y.~W. (eds.), \emph{Proceedings of the 34th International Conference on Machine Learning}, volume~70 of \emph{Proceedings of Machine Learning Research}, pp.\  166--175. PMLR, 06--11 Aug 2017.

\bibitem[Arora \& Kambhampati(2023)Arora and Kambhampati]{arora2023learning}
Arora, D. and Kambhampati, S.
\newblock Learning and leveraging verifiers to improve planning capabilities of pre-trained language models, 2023.

\bibitem[Bagaria et~al.(2021)Bagaria, Senthil, and Konidaris]{DBLP:conf/icml/BagariaS021}
Bagaria, A., Senthil, J.~K., and Konidaris, G.
\newblock Skill discovery for exploration and planning using deep skill graphs.
\newblock In Meila, M. and Zhang, T. (eds.), \emph{Proceedings of the 38th International Conference on Machine Learning, {ICML} 2021, 18-24 July 2021, Virtual Event}, volume 139 of \emph{Proceedings of Machine Learning Research}, pp.\  521--531. {PMLR}, 2021.

\bibitem[Branavan et~al.(2009)Branavan, Chen, Zettlemoyer, and Barzilay]{branavan-etal-2009-reinforcement}
Branavan, S., Chen, H., Zettlemoyer, L., and Barzilay, R.
\newblock Reinforcement learning for mapping instructions to actions.
\newblock In Su, K.-Y., Su, J., Wiebe, J., and Li, H. (eds.), \emph{Proceedings of the Joint Conference of the 47th Annual Meeting of the {ACL} and the 4th International Joint Conference on Natural Language Processing of the {AFNLP}}, pp.\  82--90, Suntec, Singapore, August 2009. Association for Computational Linguistics.

\bibitem[Chen \& Mooney(2011)Chen and Mooney]{mooney}
Chen, D.~L. and Mooney, R.
\newblock Learning to interpret natural language navigation instructions from observations.
\newblock In \emph{AAAI 2011}, 2011.

\bibitem[Chevalier{-}Boisvert et~al.(2019)Chevalier{-}Boisvert, Bahdanau, Lahlou, Willems, Saharia, Nguyen, and Bengio]{DBLP:conf/iclr/Chevalier-Boisvert19}
Chevalier{-}Boisvert, M., Bahdanau, D., Lahlou, S., Willems, L., Saharia, C., Nguyen, T.~H., and Bengio, Y.
\newblock Babyai: {A} platform to study the sample efficiency of grounded language learning.
\newblock In \emph{7th International Conference on Learning Representations, {ICLR} 2019, New Orleans, LA, USA, May 6-9, 2019}. OpenReview.net, 2019.

\bibitem[Cover \& Thomas(2001)Cover and Thomas]{DBLP:books/wi/01/CT2001}
Cover, T.~M. and Thomas, J.~A.
\newblock \emph{Elements of Information Theory}.
\newblock Wiley, 2001.

\bibitem[Fox et~al.(2017)Fox, Krishnan, Stoica, and Goldberg]{DBLP:journals/corr/FoxKSG17}
Fox, R., Krishnan, S., Stoica, I., and Goldberg, K.
\newblock Multi-level discovery of deep options.
\newblock \emph{CoRR}, abs/1703.08294, 2017.

\bibitem[Frome et~al.(2013)Frome, Corrado, Shlens, Bengio, Dean, Ranzato, and Mikolov]{NIPS2013_7cce53cf}
Frome, A., Corrado, G.~S., Shlens, J., Bengio, S., Dean, J., Ranzato, M.~A., and Mikolov, T.
\newblock Devise: A deep visual-semantic embedding model.
\newblock In Burges, C., Bottou, L., Welling, M., Ghahramani, Z., and Weinberger, K. (eds.), \emph{Advances in Neural Information Processing Systems}, volume~26. Curran Associates, Inc., 2013.

\bibitem[Fu et~al.(2023)Fu, Yu, Tiwari, Littman, and Konidaris]{DBLP:conf/icml/FuYTL023}
Fu, H., Yu, S., Tiwari, S., Littman, M., and Konidaris, G.
\newblock Meta-learning parameterized skills.
\newblock In Krause, A., Brunskill, E., Cho, K., Engelhardt, B., Sabato, S., and Scarlett, J. (eds.), \emph{International Conference on Machine Learning, {ICML} 2023, 23-29 July 2023, Honolulu, Hawaii, {USA}}, volume 202 of \emph{Proceedings of Machine Learning Research}, pp.\  10461--10481. {PMLR}, 2023.

\bibitem[Garg et~al.(2022)Garg, Vaidyanath, Kim, Song, and Ermon]{DBLP:conf/nips/GargVKSE22}
Garg, D., Vaidyanath, S., Kim, K., Song, J., and Ermon, S.
\newblock {LISA:} learning interpretable skill abstractions from language.
\newblock In Koyejo, S., Mohamed, S., Agarwal, A., Belgrave, D., Cho, K., and Oh, A. (eds.), \emph{Advances in Neural Information Processing Systems 35: Annual Conference on Neural Information Processing Systems 2022, NeurIPS 2022, New Orleans, LA, USA, November 28 - December 9, 2022}, 2022.

\bibitem[Haarnoja et~al.(2018)Haarnoja, Zhou, Abbeel, and Levine]{DBLP:conf/icml/HaarnojaZAL18}
Haarnoja, T., Zhou, A., Abbeel, P., and Levine, S.
\newblock Soft actor-critic: Off-policy maximum entropy deep reinforcement learning with a stochastic actor.
\newblock In Dy, J.~G. and Krause, A. (eds.), \emph{Proceedings of the 35th International Conference on Machine Learning, {ICML} 2018, Stockholmsm{\"{a}}ssan, Stockholm, Sweden, July 10-15, 2018}, volume~80 of \emph{Proceedings of Machine Learning Research}, pp.\  1856--1865. {PMLR}, 2018.

\bibitem[Hafner et~al.(2023)Hafner, Pasukonis, Ba, and Lillicrap]{DBLP:journals/corr/abs-2301-04104}
Hafner, D., Pasukonis, J., Ba, J., and Lillicrap, T.~P.
\newblock Mastering diverse domains through world models.
\newblock \emph{CoRR}, abs/2301.04104, 2023.

\bibitem[Hansen et~al.(2023)Hansen, Su, and Wang]{DBLP:journals/corr/abs-2310-16828}
Hansen, N., Su, H., and Wang, X.
\newblock {TD-MPC2:} scalable, robust world models for continuous control.
\newblock \emph{CoRR}, abs/2310.16828, 2023.

\bibitem[Hausman et~al.(2018)Hausman, Springenberg, Wang, Heess, and Riedmiller]{DBLP:conf/iclr/HausmanS0HR18}
Hausman, K., Springenberg, J.~T., Wang, Z., Heess, N., and Riedmiller, M.~A.
\newblock Learning an embedding space for transferable robot skills.
\newblock In \emph{6th International Conference on Learning Representations, {ICLR} 2018, Vancouver, BC, Canada, April 30 - May 3, 2018, Conference Track Proceedings}. OpenReview.net, 2018.

\bibitem[Heess et~al.(2016)Heess, Wayne, Tassa, Lillicrap, Riedmiller, and Silver]{DBLP:journals/corr/HeessWTLRS16}
Heess, N., Wayne, G., Tassa, Y., Lillicrap, T.~P., Riedmiller, M.~A., and Silver, D.
\newblock Learning and transfer of modulated locomotor controllers.
\newblock \emph{CoRR}, abs/1610.05182, 2016.

\bibitem[Hessel et~al.(2018)Hessel, Modayil, van Hasselt, Schaul, Ostrovski, Dabney, Horgan, Piot, Azar, and Silver]{DBLP:conf/aaai/HesselMHSODHPAS18}
Hessel, M., Modayil, J., van Hasselt, H., Schaul, T., Ostrovski, G., Dabney, W., Horgan, D., Piot, B., Azar, M.~G., and Silver, D.
\newblock Rainbow: Combining improvements in deep reinforcement learning.
\newblock In McIlraith, S.~A. and Weinberger, K.~Q. (eds.), \emph{Proceedings of the Thirty-Second {AAAI} Conference on Artificial Intelligence, (AAAI-18), the 30th innovative Applications of Artificial Intelligence (IAAI-18), and the 8th {AAAI} Symposium on Educational Advances in Artificial Intelligence (EAAI-18), New Orleans, Louisiana, USA, February 2-7, 2018}, pp.\  3215--3222. {AAAI} Press, 2018.

\bibitem[Huang et~al.(2022)Huang, Abbeel, Pathak, and Mordatch]{huang2022language}
Huang, W., Abbeel, P., Pathak, D., and Mordatch, I.
\newblock Language models as zero-shot planners: Extracting actionable knowledge for embodied agents.
\newblock \emph{arXiv preprint arXiv:2201.07207}, 2022.

\bibitem[Ichter et~al.(2022)Ichter, Brohan, Chebotar, Finn, Hausman, Herzog, Ho, Ibarz, Irpan, Jang, Julian, Kalashnikov, Levine, Lu, Parada, Rao, Sermanet, Toshev, Vanhoucke, Xia, Xiao, Xu, Yan, Brown, Ahn, Cortes, Sievers, Tan, Xu, Reyes, Rettinghouse, Quiambao, Pastor, Luu, Lee, Kuang, Jesmonth, Joshi, Jeffrey, Ruano, Hsu, Gopalakrishnan, David, Zeng, and Fu]{DBLP:conf/corl/IchterBCFHHHIIJ22}
Ichter, B., Brohan, A., Chebotar, Y., Finn, C., Hausman, K., Herzog, A., Ho, D., Ibarz, J., Irpan, A., Jang, E., Julian, R., Kalashnikov, D., Levine, S., Lu, Y., Parada, C., Rao, K., Sermanet, P., Toshev, A., Vanhoucke, V., Xia, F., Xiao, T., Xu, P., Yan, M., Brown, N., Ahn, M., Cortes, O., Sievers, N., Tan, C., Xu, S., Reyes, D., Rettinghouse, J., Quiambao, J., Pastor, P., Luu, L., Lee, K., Kuang, Y., Jesmonth, S., Joshi, N.~J., Jeffrey, K., Ruano, R.~J., Hsu, J., Gopalakrishnan, K., David, B., Zeng, A., and Fu, C.~K.
\newblock Do as {I} can, not as {I} say: Grounding language in robotic affordances.
\newblock In Liu, K., Kulic, D., and Ichnowski, J. (eds.), \emph{Conference on Robot Learning, CoRL 2022, 14-18 December 2022, Auckland, New Zealand}, volume 205 of \emph{Proceedings of Machine Learning Research}, pp.\  287--318. {PMLR}, 2022.

\bibitem[Jang et~al.(2017)Jang, Gu, and Poole]{DBLP:conf/iclr/JangGP17}
Jang, E., Gu, S., and Poole, B.
\newblock Categorical reparameterization with gumbel-softmax.
\newblock In \emph{5th International Conference on Learning Representations, {ICLR} 2017, Toulon, France, April 24-26, 2017, Conference Track Proceedings}. OpenReview.net, 2017.

\bibitem[Jiang et~al.(2022)Jiang, Liu, Eysenbach, Kolter, and Finn]{DBLP:conf/nips/JiangLEKF22}
Jiang, Y., Liu, E.~Z., Eysenbach, B., Kolter, J.~Z., and Finn, C.
\newblock Learning options via compression.
\newblock In Koyejo, S., Mohamed, S., Agarwal, A., Belgrave, D., Cho, K., and Oh, A. (eds.), \emph{Advances in Neural Information Processing Systems 35: Annual Conference on Neural Information Processing Systems 2022, NeurIPS 2022, New Orleans, LA, USA, November 28 - December 9, 2022}, 2022.

\bibitem[Kim et~al.(2019)Kim, Ahn, and Bengio]{DBLP:conf/nips/KimAB19}
Kim, T., Ahn, S., and Bengio, Y.
\newblock Variational temporal abstraction.
\newblock In Wallach, H.~M., Larochelle, H., Beygelzimer, A., d'Alch{\'{e}}{-}Buc, F., Fox, E.~B., and Garnett, R. (eds.), \emph{Advances in Neural Information Processing Systems 32: Annual Conference on Neural Information Processing Systems 2019, NeurIPS 2019, December 8-14, 2019, Vancouver, BC, Canada}, pp.\  11566--11575, 2019.

\bibitem[Kingma \& Welling(2014)Kingma and Welling]{DBLP:journals/corr/KingmaW13}
Kingma, D.~P. and Welling, M.
\newblock Auto-encoding variational bayes.
\newblock In Bengio, Y. and LeCun, Y. (eds.), \emph{2nd International Conference on Learning Representations, {ICLR} 2014, Banff, AB, Canada, April 14-16, 2014, Conference Track Proceedings}, 2014.

\bibitem[Kipf et~al.(2019)Kipf, Li, Dai, Zambaldi, Sanchez{-}Gonzalez, Grefenstette, Kohli, and Battaglia]{DBLP:conf/icml/KipfLDZSGKB19}
Kipf, T., Li, Y., Dai, H., Zambaldi, V.~F., Sanchez{-}Gonzalez, A., Grefenstette, E., Kohli, P., and Battaglia, P.~W.
\newblock Compile: Compositional imitation learning and execution.
\newblock In Chaudhuri, K. and Salakhutdinov, R. (eds.), \emph{Proceedings of the 36th International Conference on Machine Learning, {ICML} 2019, 9-15 June 2019, Long Beach, California, {USA}}, volume~97 of \emph{Proceedings of Machine Learning Research}, pp.\  3418--3428. {PMLR}, 2019.

\bibitem[Kolve et~al.(2017)Kolve, Mottaghi, Gordon, Zhu, Gupta, and Farhadi]{DBLP:journals/corr/abs-1712-05474}
Kolve, E., Mottaghi, R., Gordon, D., Zhu, Y., Gupta, A., and Farhadi, A.
\newblock {AI2-THOR:} an interactive 3d environment for visual {AI}.
\newblock \emph{CoRR}, abs/1712.05474, 2017.

\bibitem[Konidaris \& Barto(2009)Konidaris and Barto]{DBLP:conf/nips/KonidarisB09}
Konidaris, G.~D. and Barto, A.~G.
\newblock Skill discovery in continuous reinforcement learning domains using skill chaining.
\newblock In Bengio, Y., Schuurmans, D., Lafferty, J.~D., Williams, C. K.~I., and Culotta, A. (eds.), \emph{Advances in Neural Information Processing Systems 22: 23rd Annual Conference on Neural Information Processing Systems 2009. Proceedings of a meeting held 7-10 December 2009, Vancouver, British Columbia, Canada}, pp.\  1015--1023. Curran Associates, Inc., 2009.

\bibitem[Konidaris et~al.(2012)Konidaris, Kuindersma, Grupen, and Barto]{DBLP:journals/ijrr/KonidarisKGB12}
Konidaris, G.~D., Kuindersma, S., Grupen, R.~A., and Barto, A.~G.
\newblock Robot learning from demonstration by constructing skill trees.
\newblock \emph{Int. J. Robotics Res.}, 31\penalty0 (3):\penalty0 360--375, 2012.

\bibitem[Krishnan et~al.(2017)Krishnan, Fox, Stoica, and Goldberg]{DBLP:conf/corl/KrishnanFSG17}
Krishnan, S., Fox, R., Stoica, I., and Goldberg, K.
\newblock {DDCO:} discovery of deep continuous options for robot learning from demonstrations.
\newblock In \emph{1st Annual Conference on Robot Learning, CoRL 2017, Mountain View, California, USA, November 13-15, 2017, Proceedings}, volume~78 of \emph{Proceedings of Machine Learning Research}, pp.\  418--437. {PMLR}, 2017.

\bibitem[Lee et~al.(2022)Lee, Nachum, Yang, Lee, Freeman, Guadarrama, Fischer, Xu, Jang, Michalewski, and Mordatch]{DBLP:conf/nips/LeeNYLFGFXJMM22}
Lee, K., Nachum, O., Yang, M., Lee, L., Freeman, D., Guadarrama, S., Fischer, I., Xu, W., Jang, E., Michalewski, H., and Mordatch, I.
\newblock Multi-game decision transformers.
\newblock In Koyejo, S., Mohamed, S., Agarwal, A., Belgrave, D., Cho, K., and Oh, A. (eds.), \emph{Advances in Neural Information Processing Systems 35: Annual Conference on Neural Information Processing Systems 2022, NeurIPS 2022, New Orleans, LA, USA, November 28 - December 9, 2022}, 2022.

\bibitem[Liu et~al.(2023)Liu, Jiang, Zhang, Liu, Zhang, Biswas, and Stone]{liu2023llmp}
Liu, B., Jiang, Y., Zhang, X., Liu, Q., Zhang, S., Biswas, J., and Stone, P.
\newblock Llm+p: Empowering large language models with optimal planning proficiency, 2023.

\bibitem[Liu et~al.(2022)Liu, Yuan, C{\^o}t{\'e}, Oudeyer, and Schwing]{pmlr-v162-liu22t}
Liu, I.-J., Yuan, X., C{\^o}t{\'e}, M.-A., Oudeyer, P.-Y., and Schwing, A.
\newblock Asking for knowledge ({AFK}): Training {RL} agents to query external knowledge using language.
\newblock In Chaudhuri, K., Jegelka, S., Song, L., Szepesvari, C., Niu, G., and Sabato, S. (eds.), \emph{Proceedings of the 39th International Conference on Machine Learning}, volume 162 of \emph{Proceedings of Machine Learning Research}, pp.\  14073--14093. PMLR, 17--23 Jul 2022.

\bibitem[Meier et~al.(2011)Meier, Theodorou, Stulp, and Schaal]{DBLP:conf/iros/MeierTSS11}
Meier, F., Theodorou, E.~A., Stulp, F., and Schaal, S.
\newblock Movement segmentation using a primitive library.
\newblock In \emph{2011 {IEEE/RSJ} International Conference on Intelligent Robots and Systems, {IROS} 2011, San Francisco, CA, USA, September 25-30, 2011}, pp.\  3407--3412. {IEEE}, 2011.

\bibitem[Misra et~al.(2017)Misra, Langford, and Artzi]{Misra2017MappingIA}
Misra, D.~K., Langford, J., and Artzi, Y.
\newblock Mapping instructions and visual observations to actions with reinforcement learning.
\newblock In \emph{EMNLP}, 2017.

\bibitem[Mnih et~al.(2015)Mnih, Kavukcuoglu, Silver, Rusu, Veness, Bellemare, Graves, Riedmiller, Fidjeland, Ostrovski, Petersen, Beattie, Sadik, Antonoglou, King, Kumaran, Wierstra, Legg, and Hassabis]{DBLP:journals/nature/MnihKSRVBGRFOPB15}
Mnih, V., Kavukcuoglu, K., Silver, D., Rusu, A.~A., Veness, J., Bellemare, M.~G., Graves, A., Riedmiller, M.~A., Fidjeland, A., Ostrovski, G., Petersen, S., Beattie, C., Sadik, A., Antonoglou, I., King, H., Kumaran, D., Wierstra, D., Legg, S., and Hassabis, D.
\newblock Human-level control through deep reinforcement learning.
\newblock \emph{Nat.}, 518\penalty0 (7540):\penalty0 529--533, 2015.

\bibitem[Murali et~al.(2016)Murali, Garg, Krishnan, Pokorny, Abbeel, Darrell, and Goldberg]{DBLP:conf/icra/MuraliGKPADG16}
Murali, A., Garg, A., Krishnan, S., Pokorny, F.~T., Abbeel, P., Darrell, T., and Goldberg, K.
\newblock {TSC-DL:} unsupervised trajectory segmentation of multi-modal surgical demonstrations with deep learning.
\newblock In Kragic, D., Bicchi, A., and Luca, A.~D. (eds.), \emph{2016 {IEEE} International Conference on Robotics and Automation, {ICRA} 2016, Stockholm, Sweden, May 16-21, 2016}, pp.\  4150--4157. {IEEE}, 2016.

\bibitem[Niekum et~al.(2012)Niekum, Osentoski, Konidaris, and Barto]{DBLP:conf/iros/NiekumOKB12}
Niekum, S., Osentoski, S., Konidaris, G.~D., and Barto, A.~G.
\newblock Learning and generalization of complex tasks from unstructured demonstrations.
\newblock In \emph{2012 {IEEE/RSJ} International Conference on Intelligent Robots and Systems, {IROS} 2012, Vilamoura, Algarve, Portugal, October 7-12, 2012}, pp.\  5239--5246. {IEEE}, 2012.

\bibitem[Niekum et~al.(2013)Niekum, Chitta, Barto, Marthi, and Osentoski]{DBLP:conf/rss/NiekumCBMO13}
Niekum, S., Chitta, S., Barto, A.~G., Marthi, B., and Osentoski, S.
\newblock Incremental semantically grounded learning from demonstration.
\newblock In Newman, P., Fox, D., and Hsu, D. (eds.), \emph{Robotics: Science and Systems IX, Technische Universit{\"{a}}t Berlin, Berlin, Germany, June 24 - June 28, 2013}, 2013.

\bibitem[Pashevich et~al.(2021)Pashevich, Schmid, and Sun]{DBLP:conf/iccv/PashevichS021}
Pashevich, A., Schmid, C., and Sun, C.
\newblock Episodic transformer for vision-and-language navigation.
\newblock In \emph{2021 {IEEE/CVF} International Conference on Computer Vision, {ICCV} 2021, Montreal, QC, Canada, October 10-17, 2021}, pp.\  15922--15932. {IEEE}, 2021.

\bibitem[Pertsch et~al.(2020)Pertsch, Lee, and Lim]{DBLP:conf/corl/PertschLL20}
Pertsch, K., Lee, Y., and Lim, J.~J.
\newblock Accelerating reinforcement learning with learned skill priors.
\newblock In Kober, J., Ramos, F., and Tomlin, C.~J. (eds.), \emph{4th Conference on Robot Learning, CoRL 2020, 16-18 November 2020, Virtual Event / Cambridge, MA, {USA}}, volume 155 of \emph{Proceedings of Machine Learning Research}, pp.\  188--204. {PMLR}, 2020.

\bibitem[Raffel et~al.(2020)Raffel, Shazeer, Roberts, Lee, Narang, Matena, Zhou, Li, and Liu]{DBLP:journals/jmlr/RaffelSRLNMZLL20}
Raffel, C., Shazeer, N., Roberts, A., Lee, K., Narang, S., Matena, M., Zhou, Y., Li, W., and Liu, P.~J.
\newblock Exploring the limits of transfer learning with a unified text-to-text transformer.
\newblock \emph{J. Mach. Learn. Res.}, 21:\penalty0 140:1--140:67, 2020.

\bibitem[Rao et~al.(2022)Rao, Sadeghi, Hasenclever, Wulfmeier, Zambelli, Vezzani, Tirumala, Aytar, Merel, Heess, and Hadsell]{DBLP:conf/iclr/RaoSHWZVTAMHH22}
Rao, D., Sadeghi, F., Hasenclever, L., Wulfmeier, M., Zambelli, M., Vezzani, G., Tirumala, D., Aytar, Y., Merel, J., Heess, N., and Hadsell, R.
\newblock Learning transferable motor skills with hierarchical latent mixture policies.
\newblock In \emph{The Tenth International Conference on Learning Representations, {ICLR} 2022, Virtual Event, April 25-29, 2022}. OpenReview.net, 2022.

\bibitem[Reed et~al.(2022)Reed, Zolna, Parisotto, Colmenarejo, Novikov, Barth{-}Maron, Gimenez, Sulsky, Kay, Springenberg, Eccles, Bruce, Razavi, Edwards, Heess, Chen, Hadsell, Vinyals, Bordbar, and de~Freitas]{DBLP:journals/tmlr/ReedZPCNBGSKSEBREHCHVBF22}
Reed, S.~E., Zolna, K., Parisotto, E., Colmenarejo, S.~G., Novikov, A., Barth{-}Maron, G., Gimenez, M., Sulsky, Y., Kay, J., Springenberg, J.~T., Eccles, T., Bruce, J., Razavi, A., Edwards, A., Heess, N., Chen, Y., Hadsell, R., Vinyals, O., Bordbar, M., and de~Freitas, N.
\newblock A generalist agent.
\newblock \emph{Trans. Mach. Learn. Res.}, 2022, 2022.

\bibitem[Ren et~al.(2015)Ren, He, Girshick, and Sun]{DBLP:conf/nips/RenHGS15}
Ren, S., He, K., Girshick, R.~B., and Sun, J.
\newblock Faster {R-CNN:} towards real-time object detection with region proposal networks.
\newblock In Cortes, C., Lawrence, N.~D., Lee, D.~D., Sugiyama, M., and Garnett, R. (eds.), \emph{Advances in Neural Information Processing Systems 28: Annual Conference on Neural Information Processing Systems 2015, December 7-12, 2015, Montreal, Quebec, Canada}, pp.\  91--99, 2015.

\bibitem[Riedmiller et~al.(2018)Riedmiller, Hafner, Lampe, Neunert, Degrave, de~Wiele, Mnih, Heess, and Springenberg]{DBLP:conf/icml/RiedmillerHLNDW18}
Riedmiller, M.~A., Hafner, R., Lampe, T., Neunert, M., Degrave, J., de~Wiele, T.~V., Mnih, V., Heess, N., and Springenberg, J.~T.
\newblock Learning by playing solving sparse reward tasks from scratch.
\newblock In Dy, J.~G. and Krause, A. (eds.), \emph{Proceedings of the 35th International Conference on Machine Learning, {ICML} 2018, Stockholmsm{\"{a}}ssan, Stockholm, Sweden, July 10-15, 2018}, volume~80 of \emph{Proceedings of Machine Learning Research}, pp.\  4341--4350. {PMLR}, 2018.

\bibitem[Rissanen(1978)]{DBLP:journals/automatica/Rissanen78}
Rissanen, J.
\newblock Modeling by shortest data description.
\newblock \emph{Autom.}, 14\penalty0 (5):\penalty0 465--471, 1978.

\bibitem[Shankar \& Gupta(2020)Shankar and Gupta]{DBLP:conf/icml/Shankar020}
Shankar, T. and Gupta, A.
\newblock Learning robot skills with temporal variational inference.
\newblock In \emph{Proceedings of the 37th International Conference on Machine Learning, {ICML} 2020, 13-18 July 2020, Virtual Event}, volume 119 of \emph{Proceedings of Machine Learning Research}, pp.\  8624--8633. {PMLR}, 2020.

\bibitem[Shankar et~al.(2020)Shankar, Tulsiani, Pinto, and Gupta]{DBLP:conf/iclr/ShankarTP020}
Shankar, T., Tulsiani, S., Pinto, L., and Gupta, A.
\newblock Discovering motor programs by recomposing demonstrations.
\newblock In \emph{8th International Conference on Learning Representations, {ICLR} 2020, Addis Ababa, Ethiopia, April 26-30, 2020}. OpenReview.net, 2020.

\bibitem[Sharma et~al.(2019)Sharma, Sharma, Rhinehart, and Kitani]{DBLP:conf/iclr/SharmaSRK19}
Sharma, M., Sharma, A., Rhinehart, N., and Kitani, K.~M.
\newblock Directed-info {GAIL:} learning hierarchical policies from unsegmented demonstrations using directed information.
\newblock In \emph{7th International Conference on Learning Representations, {ICLR} 2019, New Orleans, LA, USA, May 6-9, 2019}. OpenReview.net, 2019.

\bibitem[Sharma et~al.(2022)Sharma, Torralba, and Andreas]{DBLP:conf/acl/Sharma0A22}
Sharma, P., Torralba, A., and Andreas, J.
\newblock Skill induction and planning with latent language.
\newblock In Muresan, S., Nakov, P., and Villavicencio, A. (eds.), \emph{Proceedings of the 60th Annual Meeting of the Association for Computational Linguistics (Volume 1: Long Papers), {ACL} 2022, Dublin, Ireland, May 22-27, 2022}, pp.\  1713--1726. Association for Computational Linguistics, 2022.

\bibitem[Shridhar et~al.(2020{\natexlab{a}})Shridhar, Thomason, Gordon, Bisk, Han, Mottaghi, Zettlemoyer, and Fox]{DBLP:conf/cvpr/ShridharTGBHMZF20}
Shridhar, M., Thomason, J., Gordon, D., Bisk, Y., Han, W., Mottaghi, R., Zettlemoyer, L., and Fox, D.
\newblock {ALFRED:} {A} benchmark for interpreting grounded instructions for everyday tasks.
\newblock In \emph{2020 {IEEE/CVF} Conference on Computer Vision and Pattern Recognition, {CVPR} 2020, Seattle, WA, USA, June 13-19, 2020}, pp.\  10737--10746. Computer Vision Foundation / {IEEE}, 2020{\natexlab{a}}.

\bibitem[Shridhar et~al.(2020{\natexlab{b}})Shridhar, Yuan, C{\^{o}}t{\'{e}}, Bisk, Trischler, and Hausknecht]{alfworld}
Shridhar, M., Yuan, X., C{\^{o}}t{\'{e}}, M., Bisk, Y., Trischler, A., and Hausknecht, M.~J.
\newblock Alfworld: Aligning text and embodied environments for interactive learning.
\newblock \emph{CoRR}, abs/2010.03768, 2020{\natexlab{b}}.

\bibitem[Silver et~al.(2024)Silver, Dan, Srinivas, Tenenbaum, Kaelbling, and Katz]{silver2024generalized}
Silver, T., Dan, S., Srinivas, K., Tenenbaum, J., Kaelbling, L., and Katz, M.
\newblock Generalized planning in {PDDL} domains with pretrained large language models.
\newblock In \emph{AAAI Conference on Artificial Intelligence (AAAI)}, 2024.

\bibitem[Singh et~al.(2023)Singh, Blukis, Mousavian, Goyal, Xu, Tremblay, Fox, Thomason, and Garg]{progprompt}
Singh, I., Blukis, V., Mousavian, A., Goyal, A., Xu, D., Tremblay, J., Fox, D., Thomason, J., and Garg, A.
\newblock Progprompt: Generating situated robot task plans using large language models.
\newblock In \emph{2023 IEEE International Conference on Robotics and Automation (ICRA)}, pp.\  11523--11530, 2023.
\newblock \doi{10.1109/ICRA48891.2023.10161317}.

\bibitem[Song et~al.(2023)Song, Wu, Washington, Sadler, Chao, and Su]{song2023llmplanner}
Song, C.~H., Wu, J., Washington, C., Sadler, B.~M., Chao, W.-L., and Su, Y.
\newblock Llm-planner: Few-shot grounded planning for embodied agents with large language models.
\newblock In \emph{Proceedings of the IEEE/CVF International Conference on Computer Vision (ICCV)}, October 2023.

\bibitem[Sutton et~al.(1999)Sutton, Precup, and Singh]{DBLP:journals/ai/SuttonPS99}
Sutton, R.~S., Precup, D., and Singh, S.
\newblock Between mdps and semi-mdps: {A} framework for temporal abstraction in reinforcement learning.
\newblock \emph{Artif. Intell.}, 112\penalty0 (1-2):\penalty0 181--211, 1999.

\bibitem[Tellex et~al.(2011)Tellex, Kollar, Dickerson, Walter, Banerjee, Teller, and Roy]{Tellex2011UnderstandingNL}
Tellex, S., Kollar, T., Dickerson, S., Walter, M.~R., Banerjee, A., Teller, S., and Roy, N.
\newblock Understanding natural language commands for robotic navigation and mobile manipulation.
\newblock In \emph{AAAI}, 2011.

\bibitem[Wang et~al.(2023)Wang, Xie, Jiang, Mandlekar, Xiao, Zhu, Fan, and Anandkumar]{DBLP:journals/corr/abs-2305-16291}
Wang, G., Xie, Y., Jiang, Y., Mandlekar, A., Xiao, C., Zhu, Y., Fan, L., and Anandkumar, A.
\newblock Voyager: An open-ended embodied agent with large language models.
\newblock \emph{CoRR}, abs/2305.16291, 2023.

\bibitem[Wong et~al.(2023)Wong, Mao, Sharma, Siegel, Feng, Korneev, Tenenbaum, and Andreas]{wong2023learning}
Wong, L., Mao, J., Sharma, P., Siegel, Z.~S., Feng, J., Korneev, N., Tenenbaum, J.~B., and Andreas, J.
\newblock Learning adaptive planning representations with natural language guidance, 2023.

\bibitem[Xu et~al.(2022)Xu, Veloso, and Song]{DBLP:conf/nips/XuVS22}
Xu, M., Veloso, M., and Song, S.
\newblock Aspire: Adaptive skill priors for reinforcement learning.
\newblock In Koyejo, S., Mohamed, S., Agarwal, A., Belgrave, D., Cho, K., and Oh, A. (eds.), \emph{Advances in Neural Information Processing Systems 35: Annual Conference on Neural Information Processing Systems 2022, NeurIPS 2022, New Orleans, LA, USA, November 28 - December 9, 2022}, 2022.

\bibitem[Zhang et~al.(2021)Zhang, Pertsch, Yang, and Lim]{zhang2021minimum}
Zhang, J., Pertsch, K., Yang, J., and Lim, J.~J.
\newblock Minimum description length skills for accelerated reinforcement learning.
\newblock In \emph{Self-Supervision for Reinforcement Learning Workshop-ICLR}, volume 2021, 2021.

\bibitem[Zhang et~al.(2023{\natexlab{a}})Zhang, Pertsch, Zhang, and Lim]{DBLP:journals/corr/abs-2306-11886}
Zhang, J., Pertsch, K., Zhang, J., and Lim, J.~J.
\newblock {SPRINT:} scalable policy pre-training via language instruction relabeling.
\newblock \emph{CoRR}, abs/2306.11886, 2023{\natexlab{a}}.

\bibitem[Zhang et~al.(2023{\natexlab{b}})Zhang, Zhang, Pertsch, Liu, Ren, Chang, Sun, and Lim]{DBLP:conf/corl/ZhangZPLRCSL23}
Zhang, J., Zhang, J., Pertsch, K., Liu, Z., Ren, X., Chang, M., Sun, S., and Lim, J.~J.
\newblock Bootstrap your own skills: Learning to solve new tasks with large language model guidance.
\newblock In Tan, J., Toussaint, M., and Darvish, K. (eds.), \emph{Conference on Robot Learning, CoRL 2023, 6-9 November 2023, Atlanta, GA, {USA}}, volume 229 of \emph{Proceedings of Machine Learning Research}, pp.\  302--325. {PMLR}, 2023{\natexlab{b}}.

\end{thebibliography}
\bibliographystyle{icml2024}

%%%%%%%%%%%%%%%%%%%%%%%%%%%%%%%%%%%%%%%%%%%%%%%%%%%%%%%%%%%%%%%%%%%%%%%%%%%%%%%
%%%%%%%%%%%%%%%%%%%%%%%%%%%%%%%%%%%%%%%%%%%%%%%%%%%%%%%%%%%%%%%%%%%%%%%%%%%%%%%
% APPENDIX
%%%%%%%%%%%%%%%%%%%%%%%%%%%%%%%%%%%%%%%%%%%%%%%%%%%%%%%%%%%%%%%%%%%%%%%%%%%%%%%
%%%%%%%%%%%%%%%%%%%%%%%%%%%%%%%%%%%%%%%%%%%%%%%%%%%%%%%%%%%%%%%%%%%%%%%%%%%%%%%
\newpage
\appendix
\onecolumn
\section{Example Input\&Output of Initial Segmentation with LLM}
\label{llm example}
\begin{center}
    \fcolorbox{black}{red!10}{\parbox{.9\linewidth}{
{\bf Input}:

Goal:place a microwaved apple slice on top of the black table

Action sequence: LookDown15, MoveAhead300, RotateRight90, MoveAhead100, LookDown15, PickupObject ButterKnife, SliceObject Apple, LookUp15, RotateLeft180, MoveAhead25, RotateRight90, LookDown15, PutObject DiningTable, LookUp15, RotateRight90, MoveAhead25, LookDown15, PickupObject Apple, LookUp15, RotateRight90, MoveAhead25, RotateRight90, MoveAhead150, RotateLeft90, MoveAhead250, RotateRight90, MoveAhead75, RotateLeft90, MoveAhead25, OpenObject Microwave, PutObject Microwave, CloseObject Microwave, ToggleObjectOn Microwave, ToggleObjectOff Microwave, OpenObject Microwave, PickupObject Apple, CloseObject Microwave, RotateLeft90, MoveAhead100, RotateLeft90, MoveAhead300, RotateRight90, MoveAhead75, PutObject SideTable.

{\bf Output}:
    Action alignment: 
    
    [
        \{
            index: 0,
            summary action str: navigate to kitchen step 1,
            robot actions str: LookDown 15, MoveAhead300
        \},
        
        \{
            index: 1,
            summary action str: navigate to kitchen step 2,
            robot actions str: RotateRight90, MoveAhead100, LookDown15
        \},
        
        \{
            index: 2,
            summary action str: "pick up butterknife and slice apple",
            robot actions str: "PickupObject ButterKnife, SliceObject Apple"
        \},
        
        \{
            index: 3,
            summary action str: "return butterknife to table",
            robot actions str: "LookUp15, RotateLeft180, MoveAhead25, RotateRight90, LookDown15, PutObject DiningTable"
        \},
        
        \{
            index: 4,
            summary action str: "pick up apple slice",
            robot actions str: "LookUp15, RotateRight90, MoveAhead25, LookDown15, PickupObject Apple"
        \},
        
        \{
            index: 5,
            summary action str: "navigate to microwave step 1",
            robot actions str: "LookUp15, RotateRight90, MoveAhead25, RotateRight90, MoveAhead150"
        \},
        
        \{
            index: 6,
            summary action str: "navigate to microwave step 2",
            robot actions str: "RotateLeft90, MoveAhead250, RotateRight90, MoveAhead75"
        \},
        
        \{
            index: 7,
            summary action str: "navigate to microwave step 3",
            robot actions str: "RotateLeft90, MoveAhead25"
        \},
        
        \{
            index: 8,
            summary action str: "place apple slice in microwave and turn on",
            robot actions str: "OpenObject Microwave, PutObject Microwave, CloseObject Microwave, ToggleObjectOn Microwave, ToggleObjectOff Microwave"
        \},
        
        \{
            index: 9,
            summary action str: "retrieve microwaved apple slice",
            robot actions str: "OpenObject Microwave, PickupObject Apple, CloseObject Microwave"
        \},
        
        \{
            index: 10,
            summary action str: "navigate to black table step 1",
            robot actions str: "RotateLeft90, MoveAhead100, RotateLeft90, MoveAhead300"
        \},
        
        \{
            index: 11,
            summary action str: "navigate to black table step 2",
            robot actions str: "RotateRight90, MoveAhead75"
        \},
        
        \{
            index: 12,
            summary action str: "place microwaved apple slice on black table",
            robot actions str: "PutObject SideTable"
        \}
    ],
    }}
\end{center}

\section{Prompt Example}
\label{prompt example}
\begin{center}
    \fcolorbox{black}{red!10}{\parbox{.9\linewidth}{
    You are watching a robot do household tasks. For the following task and sequence of actions taken by the robot, segment the actions into no more than 40 skills where each skill corresponds to one part of the action sequence. The answer should a python dictionary in the form of: \{(description of the first skill): (list of the actions that the robot took which correspond to the first skill), (description of the second skill): (list of the actions that the robot took which correspond to the second skill), etc.\}. The number of actions assigned to each skill should not exceed 5 but should be larger than 1. The segmentation should be as reasonable and fine-grained as possible. There should not be any leftover actions and should recover the given sequence of actions in the exact same order if we concatenate these actions in the order of the skills. 
    
    Example 1: 
    
    Input Goal: turn on light on bureau top while holding clock. 
    Input Actions: LookDown15, MoveAhead150, RotateLeft90, MoveAhead50, LookDown15, PickupObject AlarmClock, LookUp15, RotateLeft90, MoveAhead50, RotateRight90, MoveAhead75, RotateRight90, ToggleObjectOn DeskLamp
    
    Output: \{approach bureau (step 1): [LookDown15, MoveAhead150], approach bureau (step 2): RotateLeft90, MoveAhead50, LookDown15], pick up clock: [PickupObject AlarmClock], move to desk lamp (step 1): [LookUp15, RotateLeft90, MoveAhead50], move to desk lamp (step 2): [RotateRight90, MoveAhead75, RotateRight90], turn on desk lamp: [ToggleObjectOn DeskLamp]\} 

    Goal: place a microwaved apple slice on top of the black table

    Actions: LookDown15, MoveAhead300, RotateRight90, MoveAhead100, LookDown15, PickupObject ButterKnife, SliceObject Apple, LookUp15, RotateLeft180, MoveAhead25, RotateRight90, LookDown15, PutObject DiningTable, LookUp15, RotateRight90, MoveAhead25, LookDown15, PickupObject Apple, LookUp15, RotateRight90, MoveAhead25, RotateRight90, MoveAhead150, RotateLeft90, MoveAhead250, RotateRight90, MoveAhead75, RotateLeft90, MoveAhead25, OpenObject Microwave, PutObject Microwave, CloseObject Microwave, ToggleObjectOn Microwave, ToggleObjectOff Microwave, OpenObject Microwave, PickupObject Apple, CloseObject Microwave, RotateLeft90, MoveAhead100, RotateLeft90, MoveAhead300, RotateRight90, MoveAhead75, PutObject SideTable

    }}
\end{center}

\section{Implementation and Training Details}
\label{app:implementation}
\begin{figure*}[ht!]
\centering
\includegraphics[width=0.85\textwidth]
{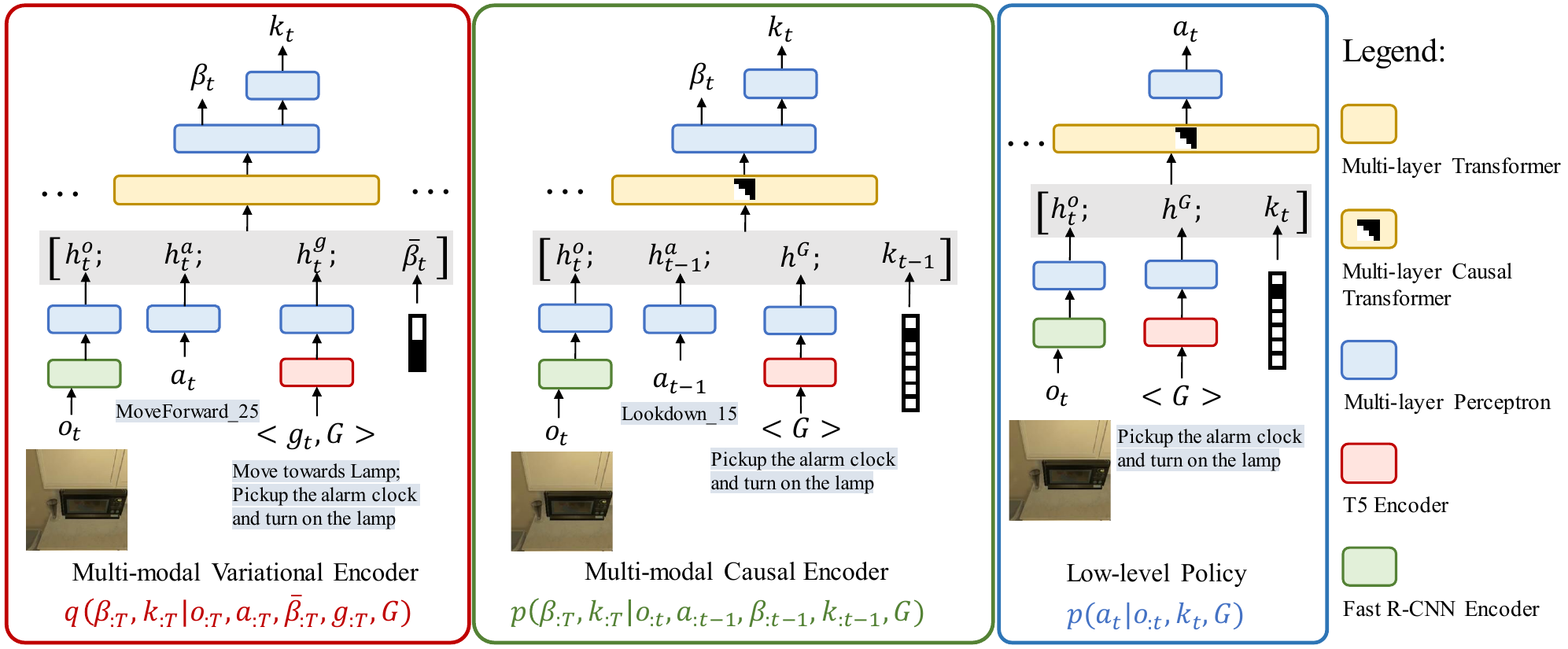}
\caption{
An overview of our Temporal Variational Inference model architecture. Each transformer corresponds to one distribution with the same color in Eqn.~\ref{eqa:obj1}.} 
\label{fig:architecture}
\end{figure*}

Note that compared with BabyAI, ALFRED is a much more complex environment and contains significantly more types of behaviors and tasks. Thus in this paper we conduct most of our downstream task evaluation and ablation studies in ALFRED.

For BabyAI, we follow the settings in the original paper. We use the partially-observable state space, with size $7\times7\times3$. We use a discrete action space with size 6: \cmd{turn left}, \cmd{turn right}, \cmd{move forward}, \cmd{pick up an object}, \cmd{drop}, \cmd{toggle}. 

\begin{figure*}[h]
\centering
\includegraphics[width=0.3\linewidth]%
{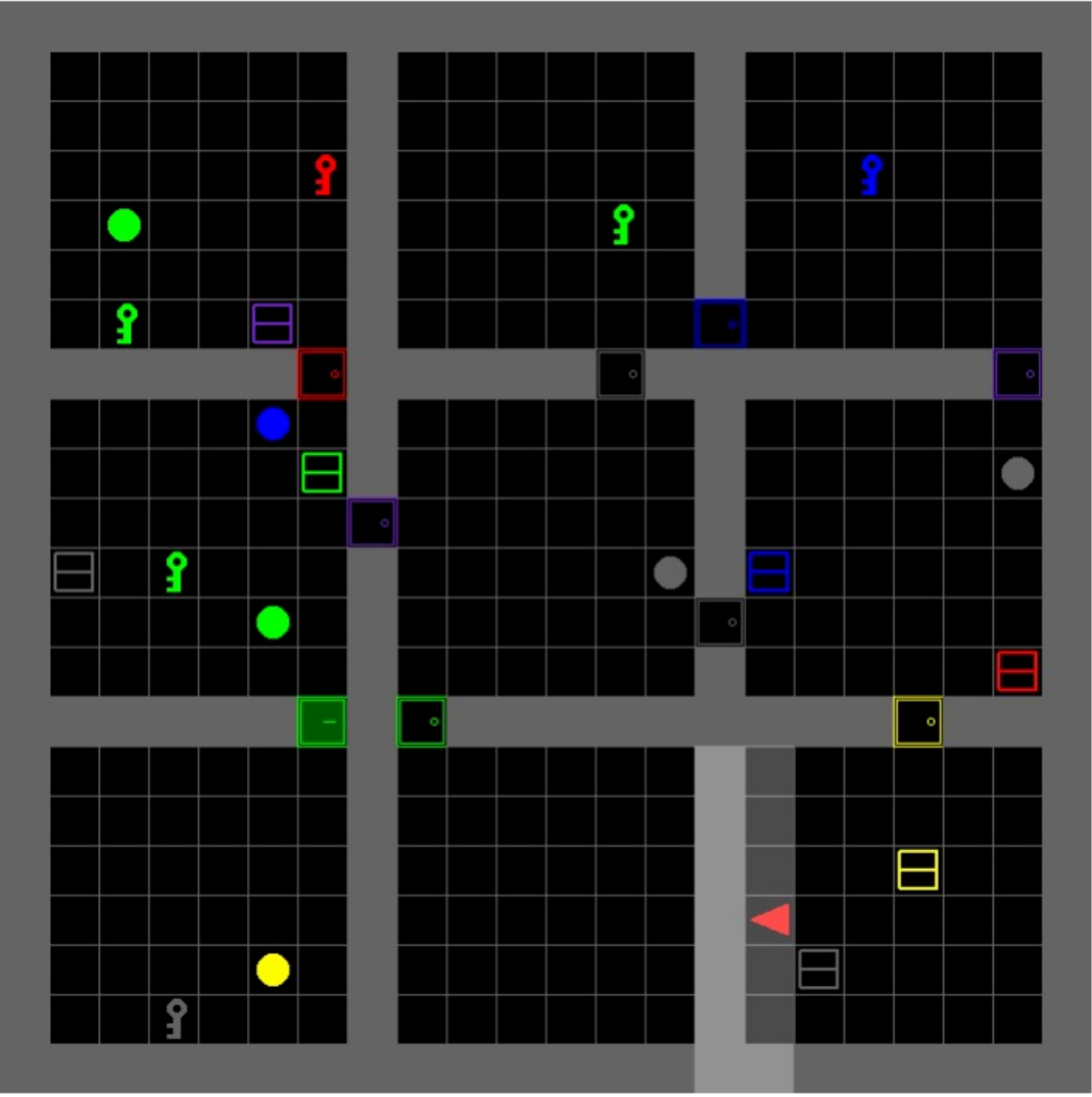}
    \caption{BabyAI} 
    \label{fig:envs}
\end{figure*}

For ALFRED, we follow the settings in~\citep{DBLP:conf/iccv/PashevichS021}. We train our algorithm on the ``training'' dataset with cross validation and test on the ``valid'' dataset. There are 6 different categories of tasks in ALFRED: \cmd{pick \& place}, \cmd{pick two \& place}, \cmd{clean \& place}, \cmd{heat \& place}, \cmd{cool \& place}, \cmd{examine in light}. And we randomly pick one task from each of them as the downstream tasks (we consider pick \& place and stack \& place as the same type of task). The action space has two components, the first component is chosen from 12 discrete action types: \cmd{Moveahead},\cmd{Lookdown}, \cmd{Lookup}, \cmd{Rotateleft}, \cmd{Rotateright}, \cmd{pick up}, \cmd{Open}, \cmd{Close}, \cmd{Put}, \cmd{Toggle on}, \cmd{Toggle off}, \cmd{slice}. The second component is chosen from 82 object types. We use the original reward function provided by the environment for downstream task training. The observations given to agents are 300$\times$ 300 images. We use a pretrained frozen Faster R-CNN encoder~\citep{DBLP:conf/nips/RenHGS15} to preprocess the image, which gives us a $512\times7\times7$ embedding input.

Also note that many recent papers for ALFRED use a planner pretrained/programmed with heuristic methods as the navigation policy, whereas in this paper all the policies are learned only from the given data with imitation learning.

For the behavior cloning (BC) baseline in ALFRED, during online training, besides the standard RL training loss (SAC), we also augment it with a KL divergence loss the keeps {\bf the prediction for the object type} close to the original prediction from the pretrained model. We empirically found that, {\bf without this auxiliary loss, BC model can hardly learn anything on the downstream tasks through the RL loss.} Note that this loss is for the object prediction only, we found that adding the same loss for the action type prediction will make the performance worse.

For our approach, across the three transformers, we have an observation encoder (two convolution layers with RELU activations, followed by one linear layer with Tanh activation) that maps from the image to an embedding with dimension 256, an action encoder (one linear layer with Tanh activation) that maps from the original action (action type + object type) to the same dimension size 256, and a language encoder (one linear layer with Tanh activation) that maps from the output of T-5 encoder to the same embedding size 256. And we concatenate all the embeddings as well as the two-dimensional boundary variable $\bar{\beta}_t$ (multi-modal variational encoder), or the one-hot vector representing the skill label $k$ (multi-modal causal encoder \& low-level policy). We use the output after passing through one linear layer as the input (512) to the transformer. For multi-modal causal encoder and low-level policy, the transformers are causally-masked transformers. The transformers have 3 layers and 8 attention heads. For multi-modal variational encoder and multi-modal causal encoder, the output is processed with one linear later, followed by another linear layer with RELU activation and gumber-softmax to output the prediction for $\beta$. Similarly, two separate linear layers with RELU activation and Gumbel-Softmax will output the prediction for $k$. For low-level policy, the output is followed by two separate MLPs with Gumbel-Softmax that outputs the predictions for action type and object type respectively. During offline training when calculating the supervised learning loss, we calculate the cross entropy loss for the prediction of action types and object types separately, and take the weighted sum of them as the total loss (weight 1 for the action type and weight 0.1 for the object type).

We provide a list of hyperparameters and their values in Table~\ref{tab:hyperparam}:

\begin{table}[h]
\centering
\caption{Hyperparameters of \ours}
\begin{tabular}{ll}
\label{tab:hyperparam}
\centering
Hyperparameters & Value \\\hline 
 % Ours (valid unseen) & 52 & 82 & 65 & 82 & 63 & 53 & 48 & 37 & 44 \\
 learning rate & $3e-4$  \\
 batch size & 16  \\
 %Ours w/o. MDL & 38 & 38 & 42 & 54 & 28 & 40 \\
 Size of skill library & 100  \\
 weight of KL loss & 0.0001 \\
 $\lambda$ & 1, 0.1, 0.01  \\
 $\gamma$ & 0.99  \\
 temperature (SAC) & 1  \\
  $\alpha_1$ & 0.01 \\
    $\alpha_2$ & 1 \\
    training epochs & 80,140\\\hline
\end{tabular}
\end{table}

\section{Temporal Variational Inference}
\label{app:tvi}
We first write the generative model for the joint distribution $p(\tau, \phi)$ as:
\begin{equation}
\label{eqa:p}
\begin{aligned}
     p(o_{:T}, a_{:T}, \beta_{:T}, g_{:T}, \bar{\beta}_{:T}, k_{:T}, G) = 
     &p(o_1) p(\bar{\beta}_{:T})p(g_{:T})p(G)
     \textstyle \\ 
     &\prod_{t=1}^{T} p(o_{t+1}\mid o_t, a_t)\pi(a_t\mid o_{:t}, k_{t}, G)
    p({\beta}_t\mid o_{:t}, a_{:t-1}, \beta_{:t-1}, k_{:t-1}, G)\\
    &p(k_t\mid o_{:t}, a_{:t-1}, k_{:t-1}, \beta_{:t}, G)
\end{aligned}
\end{equation}
where the inference for both variables $p({\beta}_t\mid o_{:t}, a_{:t-1}, \beta_{:t-1}, k_{:t-1}, G)$ and $p(k_t\mid o_{:t}, a_{:t-1}, k_{:t-1}, \beta_{:t}, G)$ is only affected by history and current variables.We add such causal constraints to ensure the inference policy can be used online where only the information till the current step can be used.

According to~\citep{DBLP:conf/icml/Shankar020}, we want to calculate the variational lower bound:
\begin{equation*}
\begin{aligned}
    \log &p(\tau) \geq \mathbb{E}_{q(\phi\mid \tau)}\log \frac{p(\tau, \phi)}{q(\phi\mid \tau)}.
\end{aligned}
\end{equation*}
Recall that in \S~\ref{sec:tvi}, we factor  $q(\phi\mid \tau)$ as:
\begin{equation*}
\begin{aligned}
    q(\beta_{:T}, k_{:T}\mid o_{:T}, a_{:T}, \bar{\beta}_{:T}, g_{:T}, G) = 
    \prod_{t=1}^{T} q(\beta_t\mid o_{:T}, a_{:T}, \bar{\beta}_t, g_{:T}, G) q(k_t\mid o_{:T}, a_{:T}, \beta_t, k_{t-1}, g_{:T}, G),
\end{aligned}
\end{equation*}
Then:
\begin{equation*}
\begin{aligned}
    \log &p(o_{:T}, a_{:T}, G) \geq \mathbb{E}_{q(\beta_{:T}, k_{:T}\mid o_{:T}, a_{:T}, \bar{\beta}_t, g_{:T}, G)}\log \frac{p(o_{:T}, a_{:T}, \bar{\beta}_{:T}, g_{:T}, \beta_{:T}, k_{:T}, G)}{q(\beta_{:T}, k_{:T}\mid o_{:T}, a_{:T}, \bar{\beta}_t, g_{:T}, G)}\\ &= \mathbb{E}_{q(\beta_{:T}, k_{:T}\mid \cdot)}[\log p(G)+\log p(\bar{\beta}_{:T}) + \log p(g_{:T}) + \sum_{t}^{T}\{\log p(o_{t+1}\mid o_t, a_t) \\ &+ \log \pi(a_t\mid o_{:t}, k_{t}, G) + \log 
    p({\beta}_t\mid o_{:t}, a_{:t-1}, \beta_{:t-1}, k_{:t-1}, G) + \log 
    p(k_t\mid o_{:t}, a_{:t-1}, k_{:t-1}, \beta_{:t}, G) \} \\ &- \log q(\beta_{:T}, k_{:T}\mid \cdot)]\\
\end{aligned}
\end{equation*}
Remove constant terms:
\begin{equation*}
\begin{aligned}
    J(\theta) &= \mathbb{E}_{q(\beta_{:T}, k_{:T}\mid \cdot)}[\sum_{t}^{T}\{\log \pi(a_t\mid o_{:t}, k_{t}, G) + \log 
    p({\beta}_t\mid o_{:t}, a_{:t-1}, \beta_{:t-1}, k_{:t-1}, G) \\ &+ \log 
    p(k_t\mid o_{:t}, a_{:t-1}, k_{:t-1}, \beta_{:t}, G) \} - \log q(\beta_{:T}, k_{:T}\mid \cdot)] \\
    &= \mathbb{E}_{q_{\theta}(\phi\mid \tau)}\sum_{t=1}^{T}\log \textcolor{darkblue}{\pi_{\theta}(a_t\mid o_{:t}, k_{t}, G)} - \sum_{t=1}^{T}\Big\{\text{KL}[\textcolor{darkred}{q_{\theta}(\beta_t\mid o_{:T}, a_{:T}, \bar{\beta}_t, g_{:T}, G)} \|  \textcolor{darkgreen}{p_{\theta}(\beta_t\mid o_{:t}, a_{:t-1}, \beta_{:t-1}, k_{:t-1}, G)}] \\ &+ \text{KL}[\textcolor{darkred}{q_{\theta}(k_t\mid o_{:T}, a_{:T}, \beta_t, k_{t-1}, g_{:T}, G)}\|  \textcolor{darkgreen}{p_{\theta}(k_t\mid o_{:t}, a_{:t-1}, k_{:t-1}, \beta_{:t}, G)}] \Big\}
\end{aligned}
\end{equation*}

\section{Minimum Description Length training objective}
\label{app:mdl}
We focus on minimizing the first term $L(\mathcal{D}\mid \phi)$.
For the original trajectory:
\begin{equation*}
    \tau = \{G, o_1, a_1, \bar{\beta}_1, g_1, o_2, a_2, \bar{\beta}_2, g_2,\cdots, o_H, a_H, \beta_H, g_H\},
\end{equation*}
we encode it as
\begin{equation*}
    \tau_{\beta, k} = \{G, o_1, k_1, \beta_1, k_2, \beta_2, \cdots, k_H, \beta_H\}.
\end{equation*}
As $\beta_t$ is a binary variable denoting whether to switch the skill at timestep $t$, we can further write the code as:
\begin{equation*}
\tau_{\beta, k} = \{G, o_1, k^{\text{new}}_1,  k^{\text{new}}_2, \cdots, k^{\text{new}}_H\},
\end{equation*}
where: 
\begin{equation*}
k^{\text{new}}_t \sim \beta_t q_{\theta}(k_t\mid o_{:T}, a_{:T}, \beta_t, g_{:T}, G) + (1-\beta_t) k_{t-1},
\end{equation*}
which we already mentioned in Eqn.~\ref{eqa:assump}.
According to the optimal code length theory~\citep{DBLP:books/wi/01/CT2001}, the expected number of bits of code generated by $p(k)$ is:
\begin{equation*}
L_k = \mathbb{E}_k[-\log p(k)].
\end{equation*}
Then for each $k^{new}_t$, the expected length is:
\begin{equation}
\begin{aligned}
&L_{k^{new}_t} =  -\mathbb{E}_k \log [q_{\theta}(k_t\mid o_{:T}, a_{:T}, \beta_t, g_{:T}, G)q_t(\beta_t=1\mid \cdot) + q_t(\beta_t=0\mid \cdot)\mathbbm{1}(k==k_{t-1})],
    \end{aligned}
\end{equation}
Then for the expected number of bits of code for the whole trajectory, as $o_1$ and $G$ are constant terms not affected by $\theta$:
\begin{equation}
\begin{aligned}
\label{eqa:appmdl}
&\mathcal{L}_{\text{MDL}}(\theta) = -\sum_t \mathbb{E}_k \log [q_t(k\mid \cdot)q_t(\beta_t=1\mid \cdot) + q_t(\beta_t=0\mid \cdot)\mathbbm{1}(k==k_{t-1})],
    \end{aligned}
\end{equation}

In LOVE~\citep{DBLP:conf/nips/JiangLEKF22}, the compression objective can be described as (using our denotations):
\begin{equation}
\label{eqa:love}
\begin{aligned}
L_{CL}(\theta) = n_s\mathcal{H}[k] = \mathbb{E}\Big[\sum_t \beta_t\log q_t(k\mid \cdot)\Big] 
    \end{aligned}
\end{equation}

We argue that our objective function is Eqn.~\ref{eqa:appmdl} can more accurately reflect the skill switches's influence on the code length and also increase trainability. The biggest differences here are that 1. The objective in Eqn.~\ref{eqa:love} ignores the effect of the skill $k_{t-1}$ chosen at last timestep to the length of the code for current timestep $t$, i.e., Eqn.~\ref{eqa:love} does not have the second term in Eqn.~\ref{eqa:appmdl}($q_t(\beta_t=0\mid \cdot)\mathbbm{1}(k==k_{t-1})$). The approximation may make huge differences in practice as shown in the following example (next paragraph). 2. In Eqn.~\ref{eqa:love} the sampled $\beta$ (binary values) are used to calculate the overall objective and then use straight-through estimator to approximate the gradients for $q_{\theta}(\beta)$, while our formulation allows us to directly calculate the gradients through $q_{\theta}(\beta)$, giving us less noisy training signals, and thus increase trainability.

Specifically, consider an example where the skill label $k$ has three possible values (1, 2, 3) at timestep $t$, $q_t(1\mid \cdot) = 0.1, q_t(2\mid \cdot) = 0.2, q_t(3\mid \cdot) = 0.7$, where choosing the third skill has the largest probability, and $q_t(\beta=1\mid \cdot) = 0.1, q_t(\beta=0\mid \cdot)=0.9$. Using Eqn.~\ref{eqa:love} in LOVE, {\bf no matter what the last skill is}, the number of bits of the code at timestep $t$ can be calculated as: $l =\mathbb{E}\Big[\beta_t\log q_t(k\mid \cdot)\Big]  =-0.1 * (0.1\log0.1 + 0.2\log0.2+0.7\log0.7) - 0.9* 1\log1 = \mathbf{0.08}$. If using our objective, when the last skill $\mathbf{k_{t-1}=1}$, the results would be: $l = -\mathbb{E}_k \log [q_t(k\mid \cdot)q_t(\beta_t=1\mid \cdot) + q_t(\beta_t=0\mid \cdot)\mathbbm{1}(k==k_{t-1})] = -((0.01+0.9)\log(0.01+0.9) + 0.02\log0.02+0.07\log0.07) = \mathbf{0.35}$. When the last skill $\mathbf{k_{t-1}=3}$, the results would be: $l = -(0.9\log0.9 + 0.02\log0.02+(0.07+0.9)\log(0.07+0.9)) = \mathbf{0.20}$, which is smaller than the case (0.2) when the last skill $k_{t-1}=1$. This is what we want the code length to express: {\bf If the current inference policy gives a high probability for the skill that is the same as the last skill ({\bf 3}), we should need fewer bits to encode current timestep and the code length should be shorter, because if it's the same skill then it is just one intermediate step of the skill we chose before and we do not need to communicate it.} And the agent will be encouraged to infer fewer boundaries (switching skills) in each trajectory. In contrast, the objective in LOVE does not reflect this property and the code length remains the same for different last skills and thus may increase the difficulty of optimization in practice.

Note that~\citet{zhang2021minimum} also leverages MDL to learn skills but their proposed objective is actually equivalent to variational inference with a different graphical model.
% \begin{figure}[htbp]
% \centering
%     \includegraphics[width=0.8\linewidth]{icml2024-1.pdf}
%     \caption{} 
%     \label{fig:app1}
% \end{figure}
% \begin{figure}[htbp]
% \centering
%     \includegraphics[width=0.8\linewidth]{icml2024-2.pdf}
%     \caption{} 
%     \label{fig:app1}
% \end{figure}
% \begin{figure}[htbp]
% \centering
%     \includegraphics[width=0.8\linewidth]{icml2024-3.pdf}
%     \caption{} 
%     \label{fig:app1}
% \end{figure}
\begin{figure*}[h]
\centering
\includegraphics[width=0.9\linewidth]%
{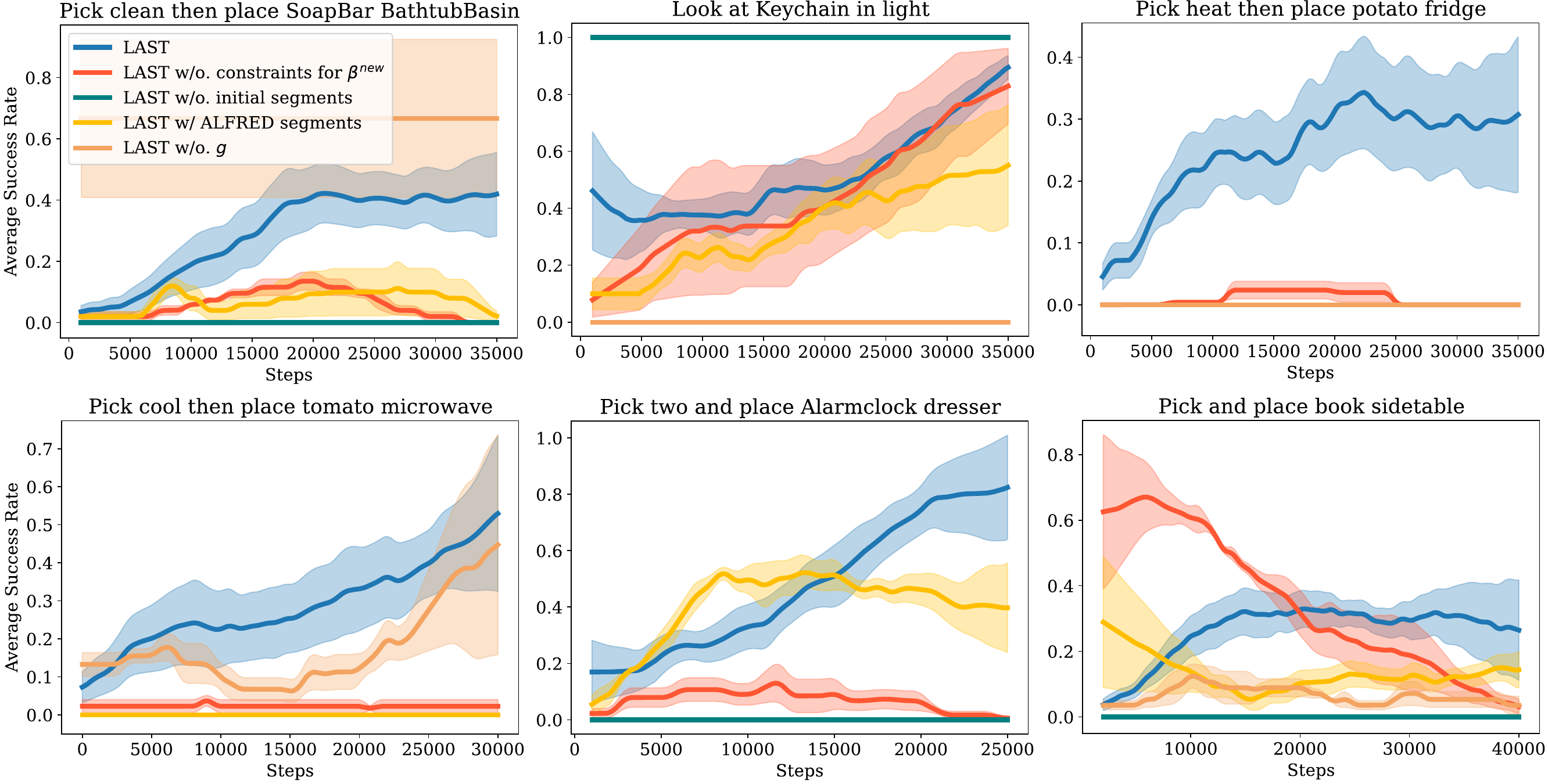}
    \caption{Ablation studies of \ours against different variants in six downstream tasks of ALFRED. We plot success rate against timesteps.} 
    \label{fig:app_abl}
\end{figure*}
\section{Ablation study and more experimental results}
\label{app:abl}
We compare \ours with several variants: 1. Using the segments and language instructions given by ALFRED dataset in place of LLM generated initial segmentation. 2. Removing the constraints on inferring $\beta^{\text{new}}_t$ (Eqn.~\ref{eqa:assump}) during temporal variational inference. 3. Removing the first step (initial segmentation) and directly do temporal variational inference training on the original data. 4. Removing the generated language annotation $g$ from the input to the variational inference model (only giving it the segmentation results $\bar{\beta}_{:T}$).

\begin{table}[h]
\footnotesize
\centering
\caption{Online (zero-shot) task success rate comparison on BabyAI.}
\begin{tabular}{lllllll}
\label{tab:appbabyaizero}
\centering
Model & \rotatebox{90}{BossLevel} & \rotatebox{90}{MiniBoss} & \rotatebox{90}{Synth} & \rotatebox{90}{SynthLoc} & \rotatebox{90}{SynthSeq} & \rotatebox{90}{Average}\\\hline 
 % Ours (valid unseen) & 52 & 82 & 65 & 82 & 63 & 53 & 48 & 37 & 44 \\
 LOVE & 27 & 31 & 40 & 63 & 25 & 37  \\
 BC & 25 & 23 & {\bf 53} & 48 & 20 & 34  \\
 %Ours w/o. MDL & 38 & 38 & 42 & 54 & 28 & 40 \\
 LISA & 27 & 26 & 49 & 49 & 27 & 36  \\
  {\bf \ours w/o. MDL} & {\bf 38} & {\bf 38} & 42 &  54 &  28 &  40 \\
    {\bf \ours} & 36 & 34 & 48 & {\bf 66} & {\bf 32} & {\bf 43} \\\hline
\end{tabular}
\end{table}
\begin{table}[h]
\small
\centering
\caption{Online (zero-shot) subtask success rate comparison for different number of skills discovered on ALFRED. The 4 \& 5 skills results are generated by setting the weight of MDL loss term to 1. The 9 \& 11 skills results are from weight 0.1, and the 15 skills \& 20 skills are from weight 0.01.}
\begin{tabular}{llllllllll}
\centering
\label{tab:numskills}
Model & \rotatebox{90}{Avg} & \rotatebox{90}{Clean} & \rotatebox{90}{Cool} & \rotatebox{90}{Heat} & \rotatebox{90}{Pick} & \rotatebox{90}{Put} & \rotatebox{90}{Slice} & \rotatebox{90}{Toggle} & \rotatebox{90}{GoTO}  \\\hline 
 4 skills & 47.0 & 82.3 & 22.0 & 9.6 & 63.8 & 48.9 & 24.3 & 46.8 & 41.8 \\
 5 skills & 48.4 & 70.8 & 34.9 & 63.2 & 62.0 & 51.0 & 40.5 & 42.2 &40.7 \\
  9 skills & 49.6 & 83.2 & 32.1 & 40.4 & 64.5 & 56.4 & 37.8 & 37.6 & 41.3 \\
  11 skills & 52.1 & 81.9 & 65.0 & 82.0 & 63.4 & 52.7 & 48.2 & 36.5 & 43.9 \\
  15 skills & 51.7 & 77.9 &75.2 & 95.6 & 63.0 & 57.4 & 34.2 & 69.4 & 39.1 \\
  20 skills & 50.0 & 82.3 & 45.0 & 96.3 & 60.1 & 54.1 & 45.0 & 56.6 & 39.7 \\\hline
\end{tabular}
\end{table}

We evaluate in the ``learning on Downstream Tasks'' setting. As shown in Fig.~\ref{fig:app_abl}, \ours achieves the best overall performance. \ours without the initial segments achieves 100 percent success rate in the \cmd{Look at Keychain in light} task but completely fails in all the other tasks, indicating that it overfits to the short horizon tasks. Without the constraints in Eqn.~\ref{eqa:assump}, the search space becomes much larger and \ours struggles to find the solution that leads to reusable skills. The segmentation of ALFRED dataset is labeled by human and we found that usually a trajectory is segmented into 4/5 pieces, which makes sense for decomposing each individual task, but it is hard to merge these relatively long subsequences into reusable skills as shown in the results. The initial segmentation made by LLMs gives a more detailed decomposition, as a result, we can more easily find reusable skills by merging these segments.  \ours without the generated language annotations $g$ generally performs not as good \ours, it reaches the highest performance in the pick clean task, but completely fails in the look at Keychain task where all other methods perform relatively well.

In Table~\ref{tab:appbabyaizero}, we compare our algorithm with a variant that does not use MDL as the auxiliary training objective in BabyAI zero-shot scenario. While \ours still achieves higher average success rate, the gap between using \& not using MDL is smaller than we saw in the online adaptation scenario. 

We also empirically investigate the influence of the weight for the MDL auxiliary objective. Interestingly, we find that changing the weight of the MDL loss can change the number of discovered skills in the same order. As shown in Table~\ref{tab:numskills}, changing the weight from 0.01 to 1 results in an increase in the number of skills from 4 to 20. The same table suggests that the zero-shot transfer performance using the discovered skills does not monotonically increase/decrease with respect to the number of the skills. A number in the middle (11 skills) gives the best performance.

\subsection{Directly using LLM to plan in ALFRED}
As we assume language-annotated low-level action space is available, it is reasonable to compare to a baseline where we use an LLM to directly output action probabilities and interact with the environment. We run this evaluation on ALFRED. We directly use GPT-4V(ision) as the LLM-based policy and run SayCan~\cite{DBLP:conf/corl/IchterBCFHHHIIJ22}-style zero shot evaluation on ALFRED. Specifically,  at each environment step, we provide GPT-4V with the current observation from the environment (image), the previously executed action sequence, the task goal, the language-annotated action space, example trajectories and ask it to predict the action. The action is further examined by a predefined function to make sure it is among the available ones at the current time step. We find that {\bf GPT-4V is not able to solve any one of the six downstream tasks we tested.}

\section{More Qualitative Results}
\begin{figure*}[htbp]
\centering
\includegraphics[width=0.88\textwidth]
{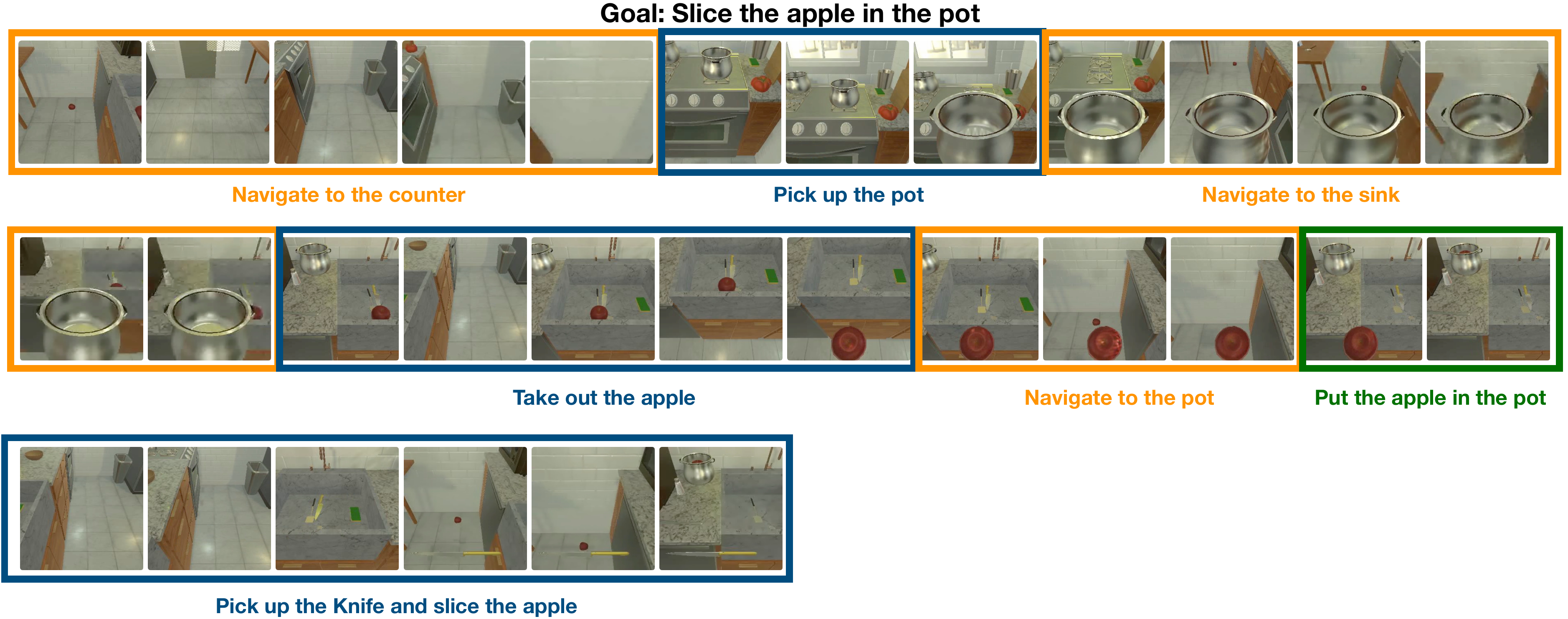}

\hspace{4cm}

\includegraphics[width=0.88\textwidth]
{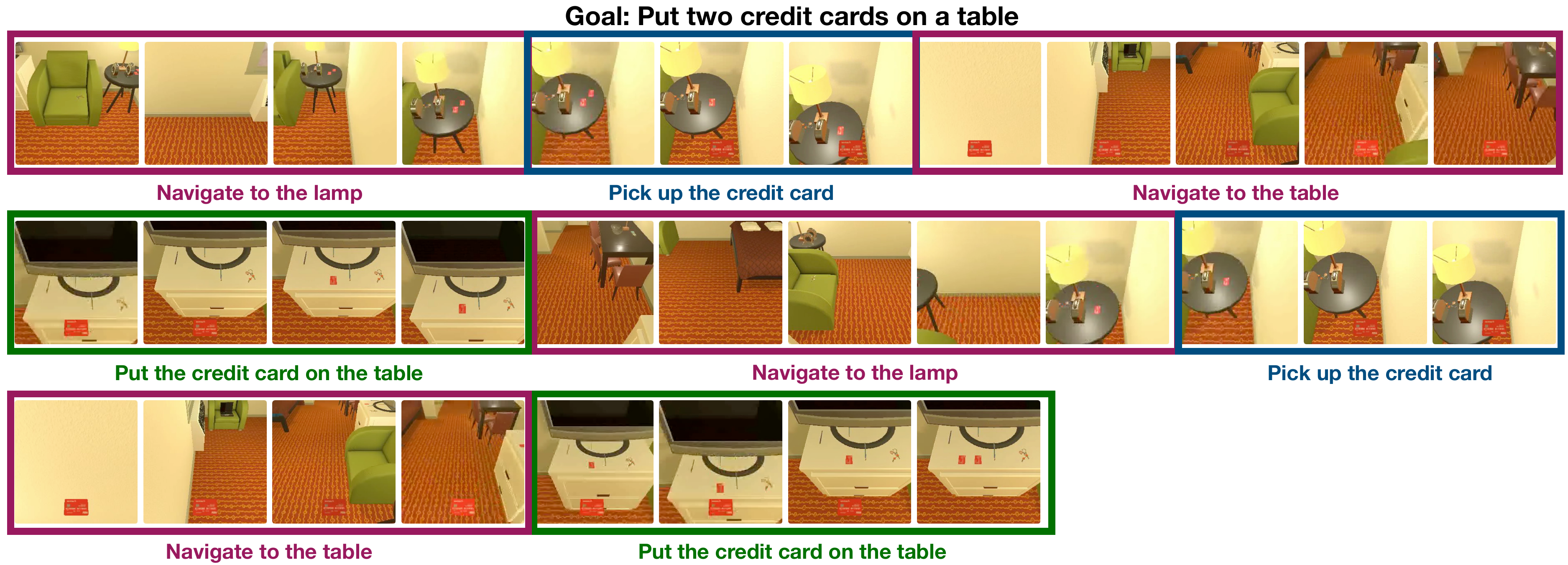}

\hspace{4cm}

\includegraphics[width=0.88\textwidth]
{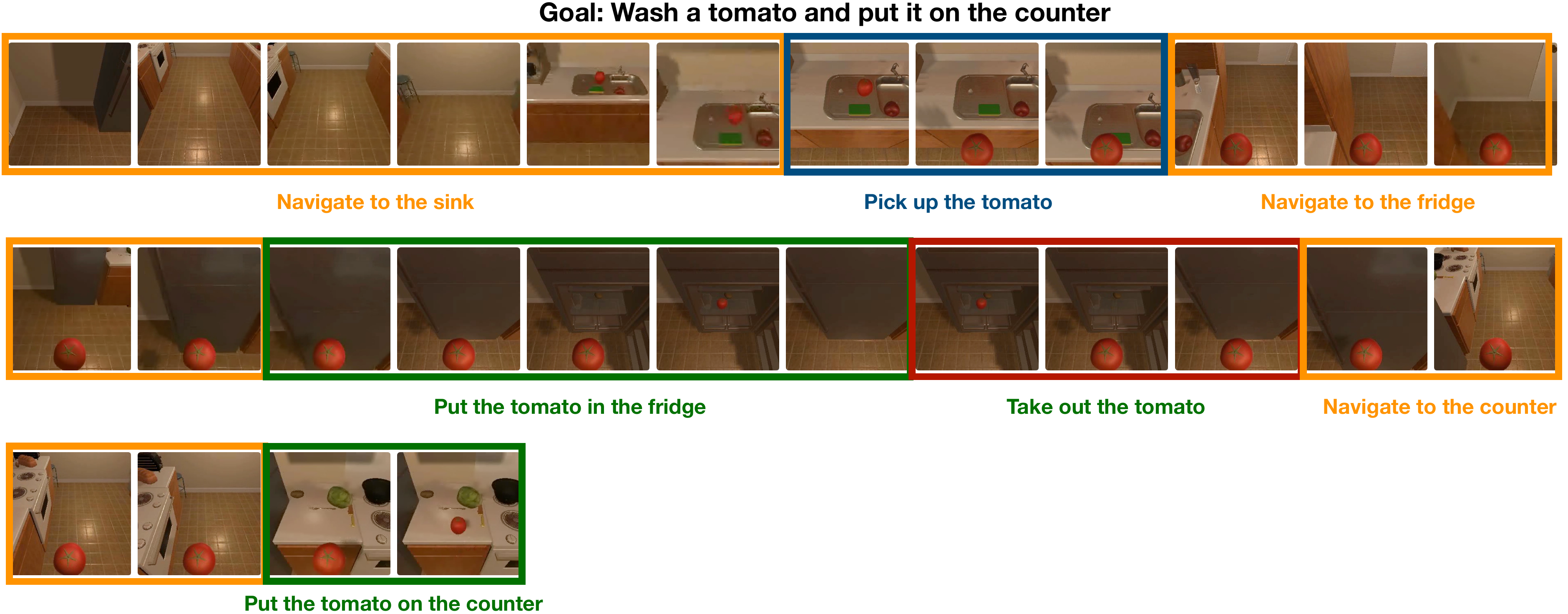}

\hspace{3cm}

\includegraphics[width=0.88\textwidth]
{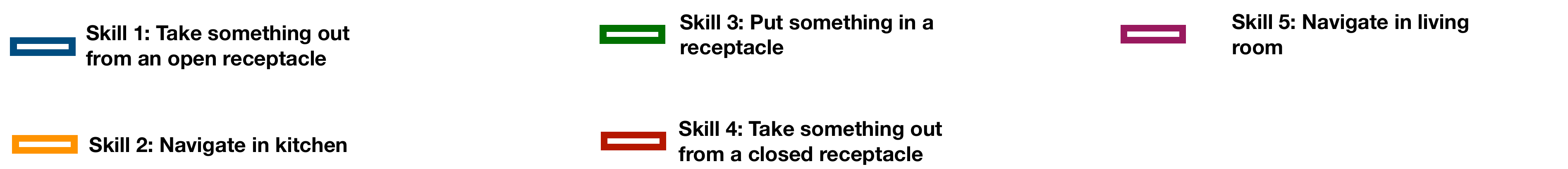}
\caption{\ours's skill segmentation for different trajectories.} 
\label{fig:appqualt}
\end{figure*}

\begin{figure*}[h]
\centering
\includegraphics[width=0.95\linewidth]%
{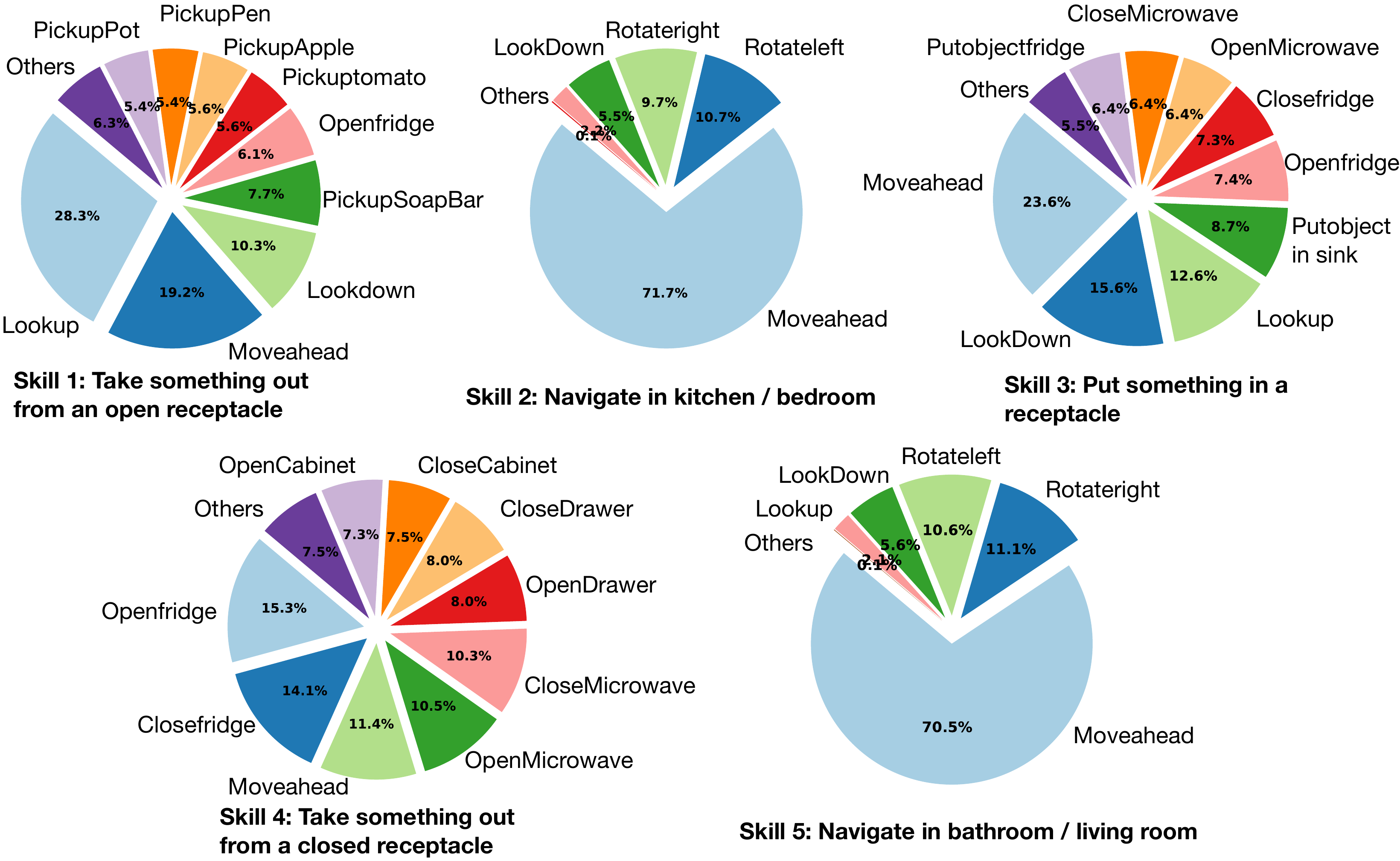}
    \caption{Five frequently discovered skills of \ours and their most-commonly used actions.} 
    \label{fig:app_pie}
\end{figure*}

We show some examples of \ours's skill segmentation in Fig.~\ref{fig:appqualt}, and the five frequently discovered skills in Fig.~\ref{fig:app_pie}.
\label{sec:appqualt}

\begin{figure*}[h]
\centering
\includegraphics[width=0.335\linewidth]%{icml2024intro.pdf}
{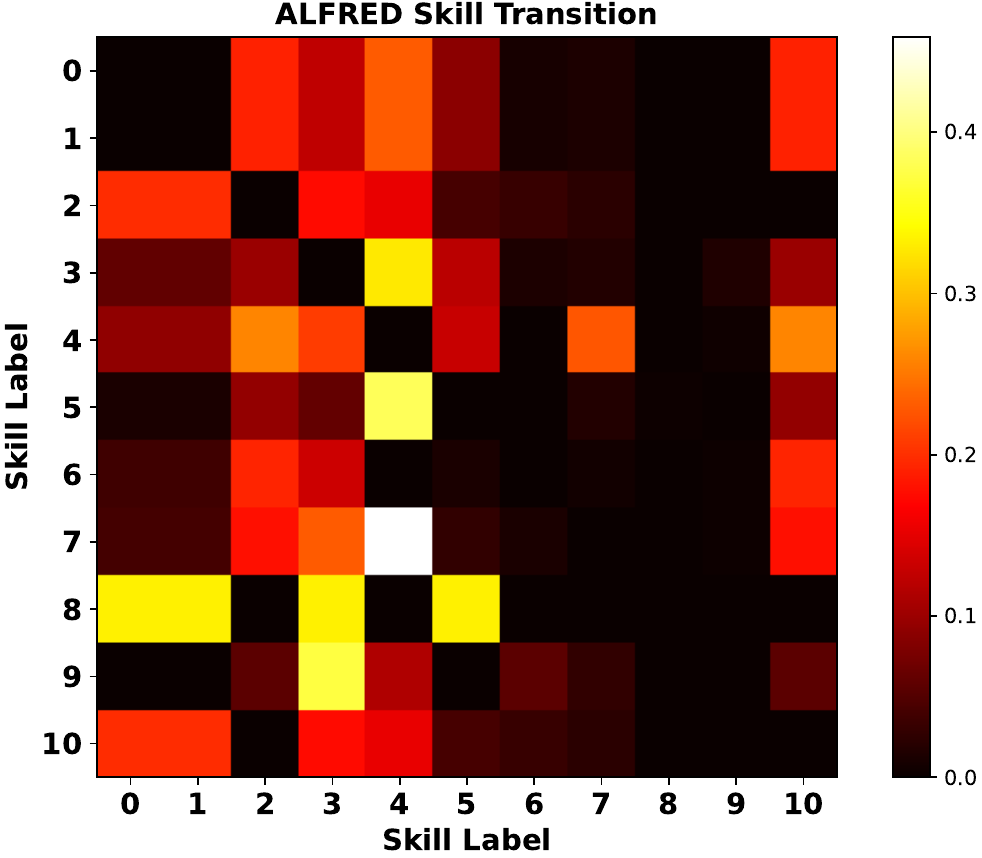}
\includegraphics[width=0.4\linewidth]%{icml2024intro.pdf}
{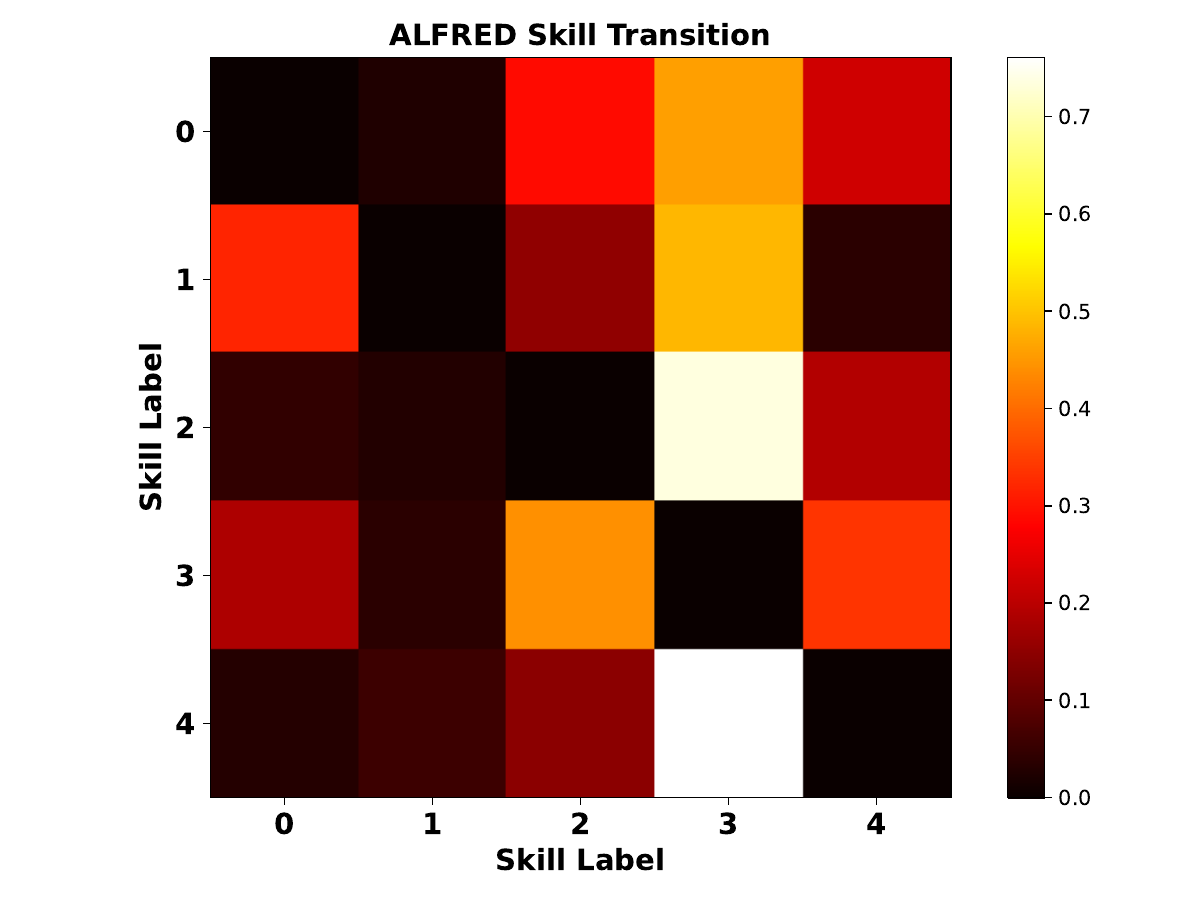}
    \caption{Transition probability matrix for two sets of learned skills from~\ours. Each row shows the probabilities of changing from one skill to all the other skills. The first five skills ($0\sim4$) correspond to the five commonly-discovered skills in Fig.~\ref{fig:qualitative}(the order may be different). We see that in the first skill set, there are another six skills discovered besides the five most common skills but the transition probabilties from the other skills to them in general are much lower than the first five skills.} 
    \label{fig:appd_trans}
\end{figure*}

\end{document}